\documentclass{article}

\newif\ifARXIV \newif\ifCR \newif\ifAPP

\ARXIVtrue \CRfalse  
\APPtrue 

\PassOptionsToPackage{numbers, compress}{natbib}

\usepackage[\ifARXIV{preprint}\else{\ifCR{final}\fi}\fi]{neurips_2022}

\usepackage{amsmath,amsfonts,bm,mathtools,amssymb}

\def\1{\bm{1}}

\def\gL{{\mathcal{L}}}

\newcommand{\R}{\mathbb{R}}
\newcommand{\D}{\mathrm{D}}

\DeclareMathOperator*{\argmin}{arg\,min}

\DeclarePairedDelimiterX{\infdivx}[2]{(}{)}{%
  #1\;\delimsize|\delimsize|\;#2%
}

\newcommand{\defeq}{\vcentcolon=}

\usepackage[utf8]{inputenc} 
\usepackage[T1]{fontenc}    
\usepackage{url}            
\usepackage{booktabs}       
\usepackage{nicefrac}       
\usepackage{microtype}      
\usepackage{xcolor}         
\usepackage{comment}
\usepackage{amsthm}
\usepackage[font=small]{caption}
\usepackage{subcaption}
\usepackage{graphicx}
\usepackage{enumerate}
\usepackage{enumitem}
\usepackage{multirow}
\usepackage{multicol}
\usepackage{wrapfig}
\usepackage{makecell}
\usepackage{pifont}
\newtheorem{theorem}{Theorem}
\newtheorem{proposition}{Proposition}

\usepackage{tikz}
\usetikzlibrary{arrows.meta, arrows, calc, matrix, backgrounds}

\usepackage{xcolor}
\definecolor{backward_arrow_color}{RGB}{168, 141, 201}
\definecolor{linkcolor}{RGB}{6,69,173}
\usepackage[
colorlinks=true,allcolors=linkcolor,pageanchor=true,
plainpages=false,pdfpagelabels,bookmarks,bookmarksnumbered,
backref=page
]{hyperref}

\usepackage[nameinlink,capitalise]{cleveref}
\Crefname{equation}{Eq.}{Eqs.}
\Crefname{section}{Sec.}{Secs.}
\Crefname{appendix}{App.}{Apps.}
\Crefname{definition}{Def.}{Defs.}
\Crefname{proposition}{Prop.}{Props.}
\Crefname{figure}{Fig.}{Figs.}

\usepackage{todonotes}

\usepackage{xspace}
\newcommand{\eg}{e.g.\xspace}
\newcommand{\ie}{i.e.\xspace}

\newcommand{\theseus}{\texttt{Theseus}\xspace}
\newcommand{\ceres}{\texttt{Ceres}\xspace}
\newcommand{\pytorch}{\texttt{PyTorch}\xspace}
\newcommand{\functorch}{\texttt{functorch}\xspace}
\newcommand{\dense}{\texttt{dense}\xspace}
\newcommand{\cusparse}{\texttt{cudaLU}\xspace}
\newcommand{\cholmodp}{\texttt{CHOLMOD}\xspace}

\newcommand{\baspacho}{\texttt{BaSpaCho}\xspace}
\newcommand{\nls}{NLS\xspace}
\newcommand{\dnls}{DNLS\xspace}
\newcommand{\implicit}{\texttt{Implicit}\xspace}
\newcommand{\unroll}{\texttt{Unroll}\xspace}
\newcommand{\trunc}{\texttt{Trunc}\xspace}
\newcommand{\truncN}[1]{\texttt{Trunc-#1}\xspace}
\newcommand{\dlm}{\texttt{DLM}\xspace}

\newcommand{\var}{\texttt{Variable}\xspace}
\newcommand{\cf}{\texttt{CostFunction}\xspace}
\newcommand{\cw}{\texttt{CostWeight}\xspace}
\newcommand{\obj}{\texttt{Objective}\xspace}
\newcommand{\opt}{\texttt{Optimizer}\xspace}
\newcommand{\adc}{\texttt{AutoDiffCostFunction}\xspace}
\newcommand{\thlayer}{\texttt{TheseusLayer}\xspace}
\newcommand{\loopbatch}{loop\textunderscore batch}

\newcommand{\cmark}{\ding{51}}%

\definecolor{luiscolor}{RGB}{50,135,168}

\definecolor{bdacolor}{RGB}{168, 141, 201}

\usepackage{listings}
\usepackage[frozencache,cachedir=.]{minted}
\usepackage{inconsolata}
\usepackage[T1]{fontenc}
\title{
Theseus:\\A Library for Differentiable Nonlinear Optimization
}

\author{
\textbf{Luis Pineda$^{1}$, 
Taosha Fan$^{1}$, 
Maurizio Monge$^{2}$, 
Shobha Venkataraman$^{1}$,}\\
\textbf{Paloma Sodhi$^{1}$, 
Ricky T. Q. Chen$^{1}$, 
Joseph Ortiz$^{1}$, 
Daniel DeTone$^{2}$, 
Austin Wang$^{1}$,}\\
\textbf{Stuart Anderson$^{1}$, 
Jing Dong$^{2}$, 
Brandon Amos$^{1}$, 
Mustafa Mukadam$^{1}$}\\[5pt]
$^{1}$Meta AI, $^{2}$Reality Labs Research
}

\begin{document}

\maketitle




\vspace{-5mm}
\begin{abstract}
\vspace{-3mm}
We present Theseus, an efficient application-agnostic open source library for differentiable nonlinear least squares (DNLS) optimization built on PyTorch, providing a common framework for end-to-end structured learning in robotics and vision. Existing DNLS implementations are application specific and do not always incorporate many ingredients important for efficiency. Theseus is application-agnostic, as we illustrate with several example applications that are built using the same underlying differentiable components, such as second-order optimizers, standard costs functions, and Lie groups. For efficiency, Theseus incorporates support for sparse solvers, automatic vectorization, batching, GPU acceleration, and gradient computation with implicit differentiation and direct loss minimization. We do extensive performance evaluation in a set of applications, demonstrating significant efficiency gains and better scalability when these features are incorporated. Project~page: \url{https://sites.google.com/view/theseus-ai/}
\end{abstract}

\vspace{-5mm}
\section{Introduction}\label{sec:intro}
\vspace{-3mm}

Reconciling traditional approaches with deep learning to leverage their complementary strengths is a common thread in a large body of recent work in robotics.
In particular, an emerging trend is to differentiate through nonlinear least squares (\nls)~\citep{nocedal2006numerical} which is a second-order optimization formulation at the heart of many problems in robotics~\citep{forster2016manifold, dellaert2017factor, Barfoot2017, mukadam2018continuous, xie2019unified, rosinol2020kimera} and vision~\citep{triggs1999bundle, pasula1999tracking, szeliski1994recovering, segal2009generalized, schmidt2014dart, schonberger2016structure}.
Optimization layers as inductive priors in neural models have been explored in machine learning with convex optimization~\citep{amos2017optnet, agrawal2019differentiable} and in meta learning with gradient descent~\citep{finn2017model, grefenstette2019generalized} based first-order optimization.

\begin{figure}[h]
    \centering
    \vspace{-3mm}
    \includegraphics[width=0.97\textwidth]{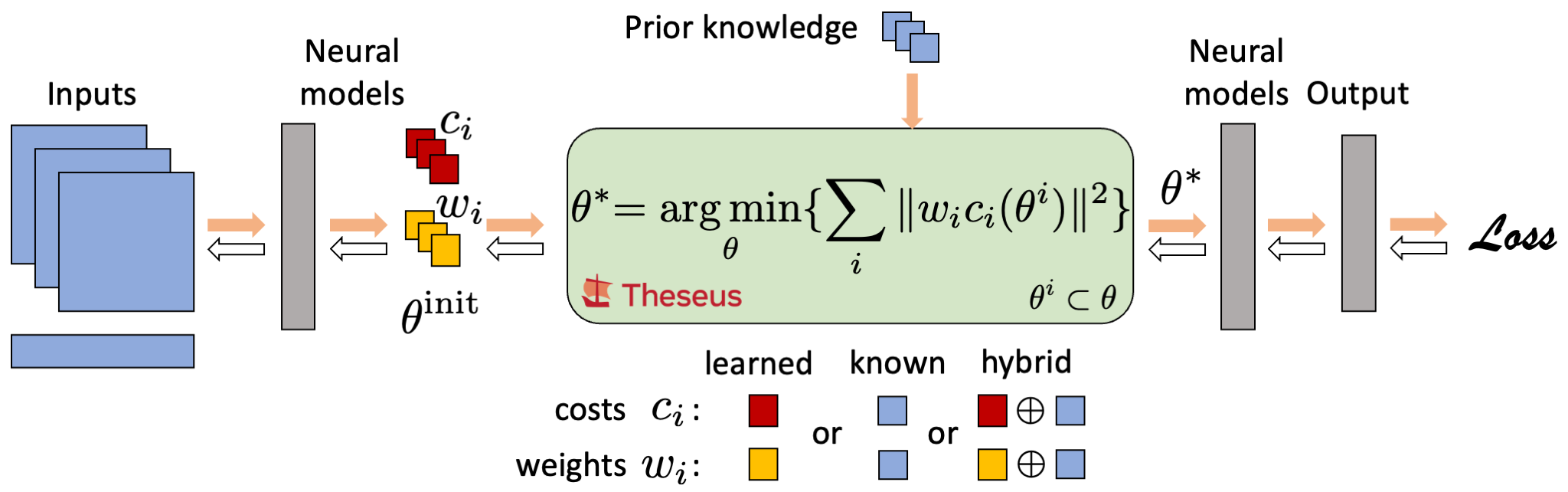}
    \caption{\theseus enables building custom, efficient \dnls layers that support end-to-end structured learning.}
    \vspace{3mm}
    \label{fig:theseuslayer}
\end{figure}

Differentiable nonlinear least squares (\dnls) provides a general scheme to encode inductive priors, as the objective function can be partly parameterized by neural models and partly with engineered domain-specific differentiable models. Here, as illustrated in \cref{fig:theseuslayer}, input tensors define a sum of weighted squares objective function and output tensors are minima of that objective. Such implicit layers~\citep{duvenaud2020deep} are in contrast to typical (explicit) layers that take input tensors through a linear transformation and some element-wise nonlinear activation function.

The ability to compute gradients end-to-end is retained by differentiating through the optimizer which allows neural models to train on the final task loss, while also taking advantage of priors captured by the optimizer. The flexibility of such a scheme has led to promising state-of-the-art results in a wide range of applications such as structure from motion~\citep{teed2018deepv2d}, motion planning~\citep{bhardwaj2020differentiable}, SLAM~\citep{jatavallabhula2020slam, teed2021droid}, bundle adjustment~\citep{Tang2019}, state estimation~\citep{yi2021differentiable, chen2022epro}, image alignment~\citep{lv2019taking} with other applications like manipulation and tactile sensing~\citep{hartley2018hybrid, sodhi2021learning}, control~\citep{giftthaler2018family}, human pose tracking~\citep{bogo2016keep, fan2021revitalizing} to be explored. However, existing implementations from above are application specific, common underlying tools like optimizers get reimplemented, and features like sparse solvers, batching, and GPU support that impact efficiency are not always included. This has led to a fragmented literature where it is difficult to start work on new ideas or to build on the progress of prior work.

To address this gap, we present \theseus, an open source library for differentiable nonlinear least squares optimization built on \pytorch. \theseus provides an efficient application-agnostic interface that consolidates recent efforts and catalyzes future progress in the domain of structured end-to-end learning for robotics and vision. Our contributions are summarized below.

\textbf{Application agnostic interface.} Our implementation provides an easy to use interface to build custom optimization layers and plug them into any neural architecture. (i) The layer can be constructed from a set of available second-order optimizers like Gauss-Newton, Levenberg–Marquardt (with adaptive damping) and Dogleg, and a nonlinear least squares objective. (ii) The objective can be constructed with learnable or hand-specified cost functions, either by applying one of many common costs already provided in the library, or by building custom costs in-place with support for automatic differentiation through \pytorch~\citep{paszke2019pytorch}. (iii) We also provide differentiable Lie groups for representing 2D/3D positions and rotations~\citep{Sola2018}, and differentiable kinematics wrapping over an existing library~\citep{meier2022differentiable} for representing robot models. More details are described in \cref{sec:api}.

\textbf{Efficiency based design.} Efficiency is a central design consideration and we make several advancements in improving computation times and memory consumption. (i) As common in prior work, an optimizer implementation using \pytorch's native linear solver would use a dense representation for solving the linear system within the nonlinear optimization. In practice, these optimization problems often have a considerable amount of sparsity that can be exploited~\citep{grisetti2011g2o, dellaert2012factor, agarwal2012ceres, Dong19ppniv}. In \theseus, we implement sparse linear solvers that are differentiable end-to-end and make them efficient with custom CPU and CUDA backends to support batching and GPU acceleration. (ii) Beyond sparse solvers, we extend batching and GPU support to all features in the library and add automatic vectorization of cost functions and other operations to significantly boost efficiency. (iii) Finally, we introduce implicit differentiation~\citep{dontchev2009implicit} and direct loss minimization~\citep{hazan2010direct, song2016training}, which have been previously applied to only first order optimizers like gradient descent and convex optimization, to a new class of second-order optimizers. This goes beyond prior work with nonlinear least squares that currently only support differentiation with standard unrolling, which is known to have challenges with compute, memory, and vanishing gradients. More details are described in \cref{sec:eff}.

\textbf{Highlights of results.} Together, the application-agnostic features let users easily set up a variety of problems like pose graph optimization, tactile state estimation, bundle adjustment, motion planning, and homography estimation, all of which are included as examples in the open source code and described in \cref{sec:applications}.
In evaluations, we find that on a standard GPU, \theseus with a sparse solver is much faster and requires significantly less memory than a dense solver, and when solving a batch of large problems the forward pass of Theseus is up to 20x faster than state-of-the-art C++ based solver \ceres that has limited GPU support and does not support batching and end-to-end learning.
We also compare all backward modes to find that with increasing number of optimization iterations, compute and memory increases linearly for unrolling and stays constant for implicit differentiation, while the latter also provides better gradients. More details are described in \cref{sec:eval}.


\vspace{-3mm}
\section{Background and related work}\label{sec:backgroud}
\vspace{-2.5mm}

\textbf{Nonlinear least squares (\nls)} is an optimization problem~\citep{nocedal2006numerical} that finds optimization variables $\theta$
\begin{equation}
  \theta^\star = \argmin_\theta S(\theta), \qquad
  S(\theta) = \frac{1}{2} \sum_{i} || r_i(\theta^i) ||^2 = \frac{1}{2} \sum_{i} || w_i c_i(\theta^i) ||^2
  \label{eq:nls}
\end{equation}
where the objective $S(\theta)$ is a sum of squared vector-valued residual terms $r_i$, each a function of $\theta^i \subset \theta$ that are (non-disjoint) subsets of the optimization variables $\theta = \{\theta_j\}$. Any variable $\theta_j$ is a manifold object; for example, a Euclidean vector or a matrix Lie group. For flexibility, we represent a residual $r_i(\theta^i) = w_i c_i(\theta^i)$ as a product of a matrix weight $w_i$ and vector cost $c_i$.
\textbf{Robotics and vision} have used this general optimization formulation to tackle many applications~\citep{dellaert2017factor, Barfoot2017}. For example, costs capture sensor measurement errors and physical constraints to optimize camera, robot, object, or human poses in estimation and tracking problems like simultaneous localization and mapping (SLAM)~\citep{dellaert2006square}, structure from motion~\citep{schonberger2016structure}, bundle adjustment~\citep{triggs1999bundle}, visual inertial odometry~\citep{forster2016manifold}, articulated tracking~\citep{schmidt2014dart}, contact odometry in legged locomotion~\citep{hartley2018hybrid}, 3D pose and shape reconstruction of humans~\citep{bogo2016keep, fan2021revitalizing} or objects~\citep{szeliski1994recovering}. Similarly, costs can also capture constraints and desired future goals to find robot states or actions in motion planning~\citep{mukadam2018continuous}, dynamics~\citep{xie2019unified}, and control~\citep{giftthaler2018family} problems.

\vspace{-0.5mm}

\textbf{Solving \nls.} Problems represented by~\cref{eq:nls} are solved by iteratively linearizing the nonlinear objective around the current variables to get the linear system $(\sum_{i} J_i^\top J_i) \delta\theta = (\sum_{i} J_i^\top r_i)$, then solving the linear system to find the update $\delta\theta$, and finally updating the variables $\theta \gets \theta - \delta\theta$, until convergence. Note that in the update the minus operation is more generally a retraction mapping for non-Euclidean variables. In the linear system, $J_i = [\partial r_i / \partial \theta^i]$ are the Jacobians of residuals with respect to the variables and the iterative method above, called Gauss-Newton (GN), is a nonlinear optimizer that is (approximately) second-order, since $H = (\sum_{i} J_i^\top J_i)$ represents the approximate Hessian. To improve robustness and convergence, variations like Levenberg–Marquardt (LM) damp the linear system, while others use a trust region and adjust step size for the update with line search (\eg, Dogleg). Please refer to~\citep{nocedal2006numerical, kelley1999iterative} for an in-depth exploration.
In most applications discussed above the objective structure gives rise to a sparse Hessian, since not all costs depend on all variables. Several general purpose frameworks~\citep{grisetti2011g2o, dellaert2012factor, agarwal2012ceres, Dong19ppniv} have been built that leverage this sparsity property to efficiently solve the sparse linear system in every iteration of the nonlinear optimization. While these frameworks were not built for deep learning, they are highly efficient and performant on CPU.

\vspace{-0.5mm}

\textbf{\nls with learning.} Data driven learning has been explored to address challenges in hand crafting costs or features for costs, finding weights to balance different costs, or to find initializations that lead to better convergence. Some examples include, learning object shape code~\citep{Sucar2020a} or environment depth code~\citep{bloesch2018codeslam} for SLAM~\citep{Czarnowski2020}, learning motion priors for planning to manipulate articulated objects~\citep{rana2017towards}, learning relative pose from tactile images to estimate object state during pushing~\citep{sodhi2021learning}, and semantic 2D segmentation fused in 3D mesh for semantic SLAM~\citep{rosinol2020kimera}. These approaches only train features on a surrogate or intermediate loss and then apply optimization at inference where the true downstream task loss is available but not utilized. To take full advantage of end-to-end learning, latest approaches thus are redesigning the optimization to be differentiable.

\vspace{-0.5mm}

\textbf{Differentiable \nls (\dnls)} solves the optimization in~\cref{eq:nls} and also provides gradients of the solution $\theta^\star$ with respect to any upstream neural model parameters $\phi$ that parameterize the objective $S(\theta; \phi)$ and in turn any costs $c_i(\theta^i; \phi)$, weights $w_i(\phi)$, or initialization for variables $\theta_{init}(\phi)$. The goal is to learn these parameters $\phi$ end-to-end with a downstream learning objective $L$ defined as a function of $\theta^\star$. This results in a bilevel optimization setup as shown in \cref{fig:theseuslayer}
\begin{equation}
  \text{inner loop:\;} \theta^\star(\phi) = \argmin_\theta S(\theta; \phi), \qquad
  \text{outer loop:\;} \phi^\star = \argmin_{\phi} L ( \theta^*(\phi) )
  \label{eq:dnls}
\end{equation}
where the inner loop is \dnls and the outer loop is gradient descent class of optimization that is standard in deep learning. The outer loop performs update $\phi \gets \phi + \delta\phi$ by computing $\delta\phi$ using gradients $\partial\theta^\star/\partial\phi$ through inner loop \dnls. Note that more generally the learning objective i.e. outer loss $L$ can also depend on other quantities like neural model parameters downstream of $\theta^*$, but we omit them here for clarity.

\vspace{-0.5mm}

\textbf{Recent works with \dnls} have outperformed optimization only or learning only methods by combining the strengths of classical methods with deep learning. For example, learning features for costs to represent depth in bundle adjustment~\citep{Tang2019} and monocular stereo~\citep{clark2018ls} where an initialization network also learns to predict depth and pose, learning cost weights like motion model weights in video to depth estimation~\citep{teed2018deepv2d}, obstacle avoidance weights in 2D motion planning from occupancy images~\citep{bhardwaj2020differentiable}, learning robust loss weights in image alignment~\citep{lv2019taking} and state-of-the-art dense SLAM~\citep{teed2021droid}, and confidence weights for feature matching to optimize camera pose~\citep{barbara2022end2end}. Other works, backpropagate reconstruction error to sensor model in a SLAM system~\citep{jatavallabhula2020slam}, solve large scale bundle adjustment on a GPU~\citep{huang2021deeplm}, and learn sensor and dynamics models for 2D visual object tracking and visual odometry~\citep{yi2021differentiable}.
These implementations however, are application specific which has led to repeated work in building \dnls where features like learnable costs and weights, Lie groups, and kinematics are not always present. Additionally, features that have a significant impact on performance, like sparsity and vectorization of costs are only considered by some~\citep{yi2021differentiable, huang2021deeplm, swiftfusion} or in the case of implicit differentiation for \nls optimization, have not yet been explored.

\vspace{-4mm}
\section{Application agnostic interface}\label{sec:api}
\vspace{-2.5mm}

Given the lack of a common and efficient framework for \dnls an important goal of \theseus is to provide an application-agnostic interface. In this section, we describe how we enable this with an easy-to-use core API, standard cost functions, and features like Lie groups and kinematics, and illustrate several examples using this interface. We discuss design for efficiency in the next section.

\begin{figure}[h]
\centering
\begin{minipage}{0.9\textwidth}
\vspace{-1mm}
\begin{minted}[fontsize=\footnotesize,frame=single,linenos]{python}
x_true, y_true, v_true = read_data() # shapes (1, N), (1, N), (1, 1)

x = th.Variable(torch.randn_like(x_true), name="x")
y = th.Variable(y_true, name="y")
v = th.Vector(1, name="v") # a manifold subclass of Variable for optim_vars

def error_fn(optim_vars, aux_vars):  # returns y - v * exp(x)
    x, y = aux_vars
    return y.tensor - optim_vars[0].tensor * torch.exp(x.tensor)
    
objective = th.Objective()
cost_function = th.AutoDiffCostFunction(
    [v], error_fn, y_true.shape[1], aux_vars=[x, y], 
    cost_weight=th.ScaleCostWeight(1.0))
objective.add(cost_function)
layer = th.TheseusLayer(th.GaussNewton(objective, max_iterations=10))

phi = torch.nn.Parameter(x_true + 0.1 * torch.ones_like(x_true))
outer_optimizer = torch.optim.Adam([phi], lr=0.001)
for epoch in range(10):
    solution, info = layer.forward(
        input_tensors={"x": phi.clone(), "v": torch.ones(1, 1)},
        optimizer_kwargs={"backward_mode": "implicit"})
    outer_loss = torch.nn.functional.mse_loss(solution["v"], v_true)
    outer_loss.backward()
    outer_optimizer.step()
\end{minted}
\vspace{-5pt}
\captionof{listing}{Simple \dnls example with \theseus, see \cref{sec:simple-example} for details.}
\label{lst:basic_differentiable}
\vspace{2mm}
\end{minipage}
\end{figure}

The core API lets users focus on describing the \dnls problem and their interaction with the outer loss $L$ and parameters $\phi$ within any broader \pytorch model, while the solution and differentiation are seamlessly taken care of under-the-hood. The basic components of the core API are described below with the help of a simple example in~\cref{lst:basic_differentiable} (see~\cref{sec:simple-example} for more details on the example):
\vspace{-0.5mm}
\begin{itemize}[leftmargin=6mm,topsep=0pt,itemsep=1pt]
\item \var: refers to either \emph{optimization variables}, $\theta$, or \emph{auxiliary variables} (those constant with respect to $S$, \eg, parameters $\phi$ or data tensors), which are named wrappers of \texttt{torch} batched tensors stored in \texttt{Variable.tensor} (lines 3-5).
\item \cf: defines costs $c_i$ (lines 12-14) and are also responsible for declaring which of its variables are optimization and which are auxiliary (lines 8-9),
\item \cw: defines weights $w_i$ associated with cost $c_i$ (line 14).
\item \obj: defines $S(\theta; \phi)$, and thus the structure of an optimization problem (lines 11, 15) by holding all cost functions and weights, and their associated variables. These are implicitly obtained when a \cf is added to the \obj, and \var names are used to infer which are shared by one or more \cf.
\item \opt: is the inner loop optimization algorithm (e.g. Gauss-Newton) that finds the solution $\theta^\star$ given objective $S$ (line 16).
\item \thlayer: encapsulates \opt and \obj, and serves as the interface between the \dnls block and other \texttt{torch} modules upstream or downstream (line 16).
\end{itemize}

The interface between the inner loop optimization and the outer loop's parameters and loss occurs via \texttt{TheseusLayer.forward} (lines 21-23). This receives as input a dictionary mapping variable names to \texttt{torch} tensors, which \theseus then uses to populate the corresponding \var with the tensor mapped to its name. With the input dictionary users can provide initial values for the optimization variables, data tensors, or current values for parameters $\phi$ before running the inner loop optimization.
The output of \texttt{forward} is another dictionary that maps variable names to tensors with their optimal values found in the inner loop (lines 21, 24); auxiliary variables are not modified during the forward pass. The output tensors can then be combined with other \texttt{torch} modules downstream to compute $L$, while maintaining the full differentiable computation graph (lines 24-26).

We currently provide Gauss-Newton, Levenberg–Marquardt (with adaptive damping), and Dogleg as nonlinear \opt for the inner loop, with the ability to easily add support for more optimizers in the future.
\cref{lst:basic_differentiable} uses \texttt{AutoDiffCostFunction} to construct an in-place \cf (line 12) which allows automatically calculating Jacobians $J_i$ with \pytorch (see \cref{app:adc}). Beyond this, in the library we include standard cost functions with analytical Jacobians broadly used in many applications, like Gaussian measurements, reprojection error, relative pose measurement, motion models, and collision costs. We also include a variety of robust loss functions, useful for example in handling outliers~\citep{barron2019general}, which can be easily integrated with \cf. Next we describe support for Lie groups and kinematics.

\textbf{Differentiable Lie groups.}
Lie groups are widely used in robotics and vision to represent 2D/3D positions and rotations~\citep{Sola2018}. Due to their non-Euclidean geometry, it is difficult to apply them to deep learning, which primarily operates with Euclidean tensors, but recently there is growing interest in making them compatible~\citep{yi2021differentiable,Peretroukhin2020, Teed2021a, Zhou2019a,Deng2021,Hansen-Estruch2021}. LieTorch~\citep{Teed2021a} generalizes automatic differentiation on the Lie group tangent space through local parameterization around the identity, but the implementation is complex since every operation requires a custom kernel. In contrast, \theseus computes common Lie group operators, e.g., the exponential and logarithm map, inverse, composition, etc., in closed form, and provides their corresponding analytical derivatives on the tangent space. Following~\citep{absil2009optimization}, we also implement a projection operator that allows us to project gradients computed by \pytorch's autodiff to the tangent space and use them to easily compute Jacobians and update Lie group variables correctly; a similar strategy has also been implemented in~\citep{jaxlie}. Additionally, our Lie group implementation includes a heuristic extension that allows using any of \pytorch's first-order optimizers on non-Euclidean manifolds with minimal code changes. All of these make it easy and straightforward to run optimization and train neural networks with Lie groups variables. More details in~\Cref{app:lie}.

\textbf{Differentiable kinematics.}
Many problems such as motion planning or state estimation on high degree of freedom robots like arms or mobile manipulators, involve computation of robot kinematics for collision avoidance or computing distance of end effector to goal. \theseus provides a differentiable implementation of forward kinematics by wrapping over Differentiable Robot Model~\cite{meier2022differentiable}, which builds a differentiable kinematics function from a standard robot model file. Gradients are computed through autodiff, while we also provide a more efficient, analytical manipulator Jacobian. This module can be used within any \cf in \theseus.

\vspace{-2mm}
\subsection{Example applications}\label{sec:applications}
\vspace{-1.5mm}

To illustrate the versatility of \theseus, we include a number of example \dnls applications below with more details in~\cref{app:examples}. Crucially, to implement these with \theseus, most of the effort is only in defining application-specific components such as data management, neural models, or custom \cf. With these defined, putting the full \dnls block together is a few lines of code to setup a \thlayer and an outer loop, similar to the simple example in~\cref{lst:basic_differentiable}.

\textbf{Pose graph optimization (PGO)} estimates poses from their noisy relative measurements~\citep{rosen2019SESync}. With \dnls we learn the radius of a Welsh robust cost function for outlier rejection, using the difference between estimated and ground truth poses as the outer loss on a synthetic dataset.




\textbf{Tactile state estimation} follows~\citep{sodhi2021learning}, which estimates 2D poses of an object pushed by a robot hand with an image-based tactile sensor~\citep{lambeta2020digit}. A neural network that predicts relative pose between hand and object from tactile images is learned end-to-end through the \thlayer.

\textbf{Bundle adjustment} is the problem of optimizing a 3D reconstruction formed by a set of camera images and a set of landmarks observed and matched across the images~\cite{agarwal2010bundle}. We learn the radius of a soft-kernel that penalizes outlier observations, using the average frame pose error as outer loss.

\textbf{Motion planning} considers a differentiable version of the GPMP2 planning algorithm, inspired by~\cite{bhardwaj2020differentiable}, where the outer loss tries to match expert demonstrations. Here we learn a model for initializing optimization variables, and we include the inner loop objective as a term in the outer loss.

\textbf{Homography estimation.} Homography is a linear transformation between corresponding points in two images and can be solved by minimising a dense photometric loss. Robustness to lighting and viewpoint change can be improved with a feature-metric loss based on CNN features~\cite{czarnowski2017, dong18learning,tang2018ba,sarlin21pixloc,Lv19cvpr,Wang2018DeepLKFE}. In our outer loop, we train a CNN to produce robust features for image alignment.

\vspace{-4mm}
\section{Efficiency based design}\label{sec:eff}
\vspace{-3mm}

\theseus enables several different applications with a general interface. Compute and memory efficiency are central to making its usage practical. Next, we explain design considerations to support batching and vectorization, sparsity, and backward modes for differentiation, which we demonstrate boost performance in the evaluations section.

\begin{wrapfigure}{r}{0.5\textwidth}
    \centering
    \vspace{-13.5mm}
    \includegraphics[width=0.48\textwidth]{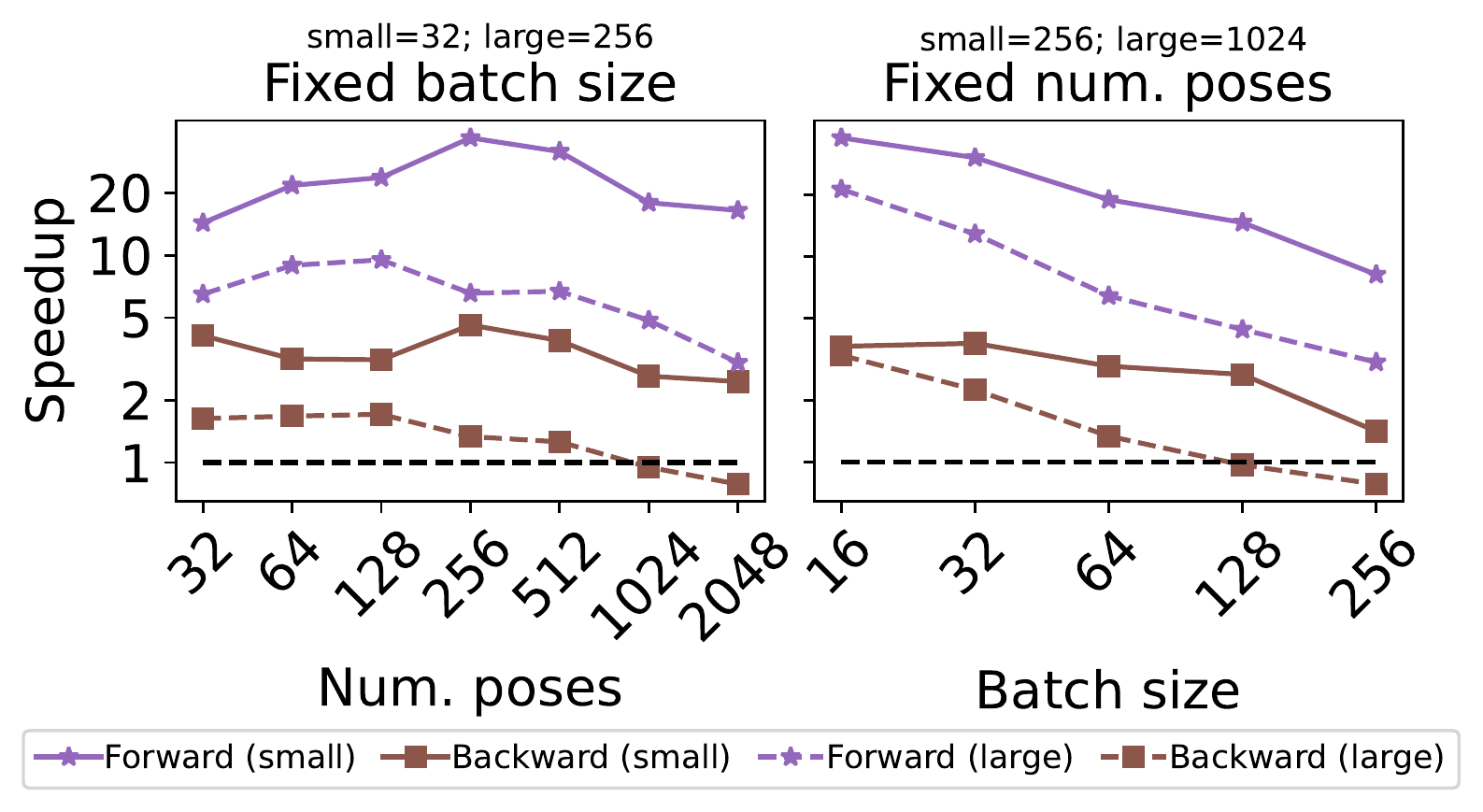}
    \vspace{-1mm}
    \caption{Speedup with automatic vectorization on PGO. Black dotted line is without vectorization.}
    \label{fig:vectorization}
\end{wrapfigure}

\vspace{-3mm}
\subsection{Batching and vectorization}
\vspace{-2mm}

Parallel processing is important to improve computational efficiency in machine learning and optimization. In \theseus, we enable two levels of parallelization. First, \theseus natively supports solving a batch of \dnls in parallel, thus fitting seamlessly in the \pytorch framework, where training and inferences on batches is the standard.
Second, inspired by DeepLM~\citep{huang2021deeplm}, and noting that lots of the operations such as costs, gradients/Jacobian computation, and variable updates only differ from each other in terms of the input data, we make use of the single-instruction-multiple-data (SIMD) protocol by automatically detecting and vectorizing operations of the same type, significantly reducing computation overhead. Using the PGO example, \cref{fig:vectorization} shows that \theseus achieves significant speedup with automatic vectorization both for forward and backward pass. Note that there is an application-dependent trade-off between memory and speed; here the memory use increases by up to $\sim82\%$ for forward and $\sim55\%$ for backward.

\vspace{-3mm}
\subsection{Handling sparsity with linear solvers beyond \pytorch}\label{sec:sparse-solver}
\vspace{-2mm}

Solving \nls requires solving a sequence of linear systems to obtain descent directions. As discussed in~\cref{sec:backgroud}, these systems are generally sparse and can be solved much more efficiently if not treated as dense. \theseus includes differentiable sparse solvers that take advantage of the sparsity, complementing \pytorch's native dense solvers. Importantly, \theseus seamlessly takes care of assembling the cost functions and variables in the objective into sparse data structures that our linear solvers can consume, without any extra burden on the user.
Currently, we provide three sparse solvers: (i) a CPU-based solver that relies on \texttt{CHOLMOD}~\cite{chen2008algorithm}, (ii) \texttt{cudaLU}, which is based on the \texttt{cuSolverRF} package that is part of Nvidia's \texttt{cuSolver} library provided with CUDA, and (iii) \baspacho, our novel batched sparse Cholesky solver with GPU support. As a bonus feature, we provide access to these solvers as standalone \pytorch functions, so they can be used to solve sparse matrices arising outside of \nls or \dnls optimization.

\textbf{\texttt{CHOLMOD}-based solver.}
\texttt{CHOLMOD}~\cite{chen2008algorithm} achieves state-of-the-art performance on computation of the Cholesky decomposition of sparse matrices. It exploits parallelism by grouping sparse entries to take advantage of high-performance multi-threaded dense matrix operations in \texttt{BLAS}/\texttt{LAPACK} libraries. \texttt{CHOLMOD} has some limited support for GPU for some of its operations, but the algorithm is strongly CPU-based, and the user is expected to provide matrix data on the CPU. One convenient feature is computing the symbolic analysis of a sparse matrix pattern as a separate step and creating a symbolic decomposition object that can be used for all subsequent factorizations. We also take advantage of builtin functionality for sparse multiplication and only provide the Jacobian matrix $J$ to solve for the Hessian $H=J^\top J$. Two limitations of the library with respect to \theseus are, first, the lack of proper GPU support, which forces us to provide matrix data on the CPU, and, second, the lack of batching, which requires us to loop to solve every problem in the batch independently. On the other hand, since it runs on CPU, it has less memory restrictions than GPU-based solvers (see~\cref{sec:profiling}). 

\textbf{\texttt{cudaLU} solver.}
\texttt{cuSolverRF} is designed to accelerate the solution of sets of linear systems by fast LU refactorization when given new coefficients for the same sparsity pattern. To take advantage of this, we implemented custom CUDA kernels for batched sparse matrix-matrix and matrix-vector products, and for solving a batch of sparse linear systems using LU factorization from \texttt{cuSolverRF}. Although this solver leads to a substantial performance boost over \pytorch's dense solver (see~\Cref{sec:profiling}), the closed-source nature of \texttt{cuSOLVER} results in some challenges and limitations:
(i) \texttt{cuSolverRF} does not support separate symbolic decomposition and numeric contexts, so it's not possible to use the same symbolic decomposition to hold in memory separate factors. Since this is necessary in \theseus for unrolling of the inner loop, we work around this limitation by creating a pool of contexts, and we use the least recently used context for factorization. As a consequence, the number of contexts must be set according to the number of iterations that need to be unrolled;
(ii) The batch size is fixed once a context is created. Since recreating the contexts is an expensive operation, it means that the batch size has to be constant over the course of outer loop optimization;
(iii) It relies on LU factorization, which for symmetric matrices (the case of \theseus) is less efficient than using Cholesky decomposition.

\textbf{\baspacho solver.} Batched Sparse Cholesky (\baspacho) is a novel open-source sparse Cholesky solver designed for \theseus with support for batching (\url{https://github.com/facebookresearch/baspacho}). \baspacho implements the \emph{supernodal} Cholesky algorithm~\citep{ng1993supernodal} to achieve state-of-the art performance by exploiting dense operations via \texttt{BLAS}/\texttt{cuBLAS}. This is achieved by building an elimination tree and then clustering column blocks with similar sparsity patterns. These blocks form nodes of the elimination tree and allow dense operations. In \baspacho, the dense operations are dispatched to \texttt{BLAS} (on CPU) or \texttt{cuBLAS} (on GPU), with additional support added on top for batching matrix operations with the same sparsity patterns. In problems with very sparse matrices, like bundle adjustment~\citep{triggs1999bundle}, the supernodal algorithm employed in state-of-the-art solvers~\citep{agarwal2012ceres} is unable to eliminate columns of parameter blocks simultaneously. Thus, past work has resorted to the Schur complement trick~\citep{zhang2006schur} to send a reduced problem to the sparse solver. However, this logic adds extra complexity to the nonlinear optimization, while essentially duplicating the work of the (mathematically equivalent) Cholesky decomposition. In \baspacho, we instead complement the supernodal algorithm with \emph{sparse elimination} that removes the need to externally handle Schur complement as a workaround to the limitation of the supernodal algorithm. More details are described in \cref{app:baspacho}.

\textbf{Backward for custom linear solvers.}
Obtaining derivatives of the linear system solve with respect to the parameters is a crucial operation for DNLS. In particular, we consider optimizing the parameters $A$ and $b$ of a linear system $y=A^{-1}b$ to minimize a downstream function $f(y)$. The derivatives of the loss with respect to the parameters of the linear system can be obtained with implicit differentiation, $\frac{\partial f}{\partial b} = A^{-1} \frac{\partial f}{\partial y}$ and $\frac{\partial f}{\partial A} = -A^{-1} \frac{\partial f}{\partial y} y^\top$, as done in~\citet{barron2016fast}.
In \theseus, we implement this by connecting the Python interface of our sparse solvers with \pytorch's \texttt{autograd.Function} classes that implement the gradients above in their \texttt{backward} methods. This connects the computation graph between the downstream function and any upstream parameters that modify the system via auxiliary variables or values for optimization variables. Furthermore, since the gradients require solving linear systems that use the same matrix as the forward pass, our backward pass can cache factorizations, resulting in it being significantly faster than the forward pass (see~\Cref{fig:profiling}).



\vspace{-3mm}
\subsection{Backward modes for \dnls}\label{sec:adjoint-differentiation}
\vspace{-2mm}

The parameters $\phi$ upstream of \dnls can be learned end-to-end through the solution $\theta^\star(\phi)$ by using the \emph{adjoint derivatives} $\partial \theta^\star(\phi)/ \partial \phi$. We include four methods for computing them in \theseus.

\textbf{Unrolling}
is the standard way in which past work in \dnls has computed the adjoint derivatives. This is often referred to as backpropagation through time or unrolled optimization and is explored in \citep{finn2017model, bhardwaj2020differentiable, pearlmutter2008reverse, zhang2010multi, maclaurin2015gradient, belanger2016structured, metz2016unrolled, han2017alternating, belanger2017end, belanger2017deep, foerster2017learning, monga2021algorithm}. In practice, often only a few steps of unrolling are performed due to challenges with compute, memory, and vanishing gradients.

\textbf{Truncated differentiation.}
Aside from unrolling a few steps, another way of approximating the derivatives is to use truncated backpropagation through time (TBPTT)~\citep{werbos1990backpropagation, jaeger2002tutorial}. Truncation unfortunately results in biased derivatives and many works~\citep{tallec2017unbiasing, wu2018understanding, liao2018reviving, shaban2019truncated, vicol2021unbiased} seek to further theoretically understand the properties of TBPTT, including the bias of the estimator and how to unbias it.

\textbf{Implicit differentiation.}
If $\theta^\star$ can be computed exactly, then the implicit function theorem provides a way of computing the adjoint derivatives, as done in related work in convex optimization and first-order gradient descent methods~\citep{amos2017optnet, agrawal2019differentiable, amos2019differentiable, domke2012generic, gould2016differentiating, lorraine2020optimizing, blondel2021efficient, atzmon2019levelsets}. We apply the implicit function theorem from~\citet[Theorem 1B.1]{dontchev2009implicit} (see~\cref{app:backward}) to~\cref{eq:dnls} to perform implicit differentiation on a new class of second-order \nls optimization. This first requires that we transform~\cref{eq:dnls} into an implicit function that finds the roots. We do this via the first-order optimality condition, resulting in $g(\theta; \phi)\defeq \nabla_\theta S(\theta; \phi)$. Finding $\Theta^\star(\phi)\defeq\{\theta \mid g(\theta;\phi)=0\}$ corresponds to solving~\cref{eq:dnls}. Under mild assumptions, the theorem above gives the adjoint derivative at $\bar \phi$
\begin{equation}
  \D_\phi \theta^\star(\bar \phi) = -\D_\theta^{-1} g(\theta^\star(\bar \phi); \bar\phi) \D_\phi g(\theta^\star(\bar\phi); \bar\phi).
  \label{eq:implicit-derivative}
\end{equation}
As \theseus internally uses a (Gauss-)Newton solver, the following proposition provided in~\cref{app:backward} shows how we can compute~\cref{eq:implicit-derivative} by differentiating a single Newton step at an optimal solution.
\begin{proposition}
\vspace{-0.5mm}
The implicit derivative (\cref{eq:implicit-derivative}) can be
computed by differentiating a Newton step
$h(\theta; \phi):=\theta-[\nabla^2_\theta S(\theta; \phi)]^{-1}_{\rm stop}\nabla_\theta S(\theta; \phi)$
at an optimal $\theta^\star$, where
$[\cdot]_{\rm stop}$ zeros the derivative.
\label{prop:newton-ift}
\vspace*{-2mm}
\end{proposition}

\textbf{Direct loss minimization.}
Suppose we have an outer loss as in~\cref{eq:dnls}. The direct loss minimization (DLM) approach uses this loss to augment the inner-loop optimization problem in order to define a finite difference scheme that approaches the true gradient
$\nabla_\phi L = \lim_{\varepsilon \rightarrow 0} g_\text{DLM}^\varepsilon$,
where
$g_\text{DLM}^\varepsilon \triangleq \tfrac{1}{\varepsilon} \{ \tfrac{\partial}{\partial \phi} S(\theta^*; \phi) - \tfrac{\partial}{\partial \phi} S(\theta_\text{direct}; \phi) \}$.
This was used in prior works that solve optimization problems on structured discrete domains \citep{hazan2010direct, song2016training,vlastelica2019differentiation,niepert2021implicit}, but has so far not seen much use in structured continuous settings. We modify the original DLM formulation to better suit its implementation within \theseus
\begin{equation}
    \theta^\star = \argmin_{\hat\theta} S(\hat\theta; \phi),
    \qquad
    \theta_\text{direct} = \argmin_{\hat\theta} S(\hat\theta; \phi) + \big\| \varepsilon \hat\theta - \tfrac{1}{2} \nabla_\theta L(\theta^*) \big\|^2.
\end{equation}
This is different from the original formulation in two ways: (i) we only assume access to the gradient vector $\nabla_\theta L(\theta^*)$, which helps formulate DLM as an algorithm for computing vector-Jacobian products, and (ii) we add a small regularization term to ensure the modified objective for $\theta_\text{direct}$ is a sum-of-squares without affecting the limit as $\varepsilon \rightarrow 0$. See \cref{app:backward} for more details.

\vspace{-3mm}
\section{Evaluation}\label{sec:eval}
\vspace{-2mm}


We evaluate the performance of \theseus under different settings with PGO and tactile state estimation applications from~\Cref{sec:applications}. PGO allows us to easily control the problem scales for performance evaluation; in~\Cref{sec:profiling} we profile time and memory consumption of \theseus in an end-to-end setup and in~\Cref{sec:ceres-comp} we evaluate timings of \theseus as a stand-alone \nls optimizer and compare with state-of-the-art \ceres~\citep{agarwal2012ceres}. The tactile state estimation application involves a more complex outer loop model that is useful for comparing all different backward modes, which we present in~\Cref{sec:backward-experiments}.

\vspace{-2mm}
\subsection{Profiling forward and backward pass of \theseus for DNLS}\label{sec:profiling}
\vspace{-2mm}

We study the performance of \theseus for DNLS on the PGO problem~\citep{rosen2019SESync} with the synthetic Cube dataset, as described in \Cref{app:examples}. We run 10 inner loop iterations and 20 outer loop epochs, and use implicit differentiation to compute gradients of the inner \nls optimization. For these experiments we used an Nvidia V100 GPU with 32GBs of memory for all Python computation, and Intel Xeon 2.2GHz CPU with 20 threads for the CPU-based \texttt{CHOLMOD} linear solver. We evaluate performance using our sparse solvers in \theseus and using \pytorch's Cholesky \dense solver.

\begin{figure}
    \centering
    \includegraphics[width=\textwidth]{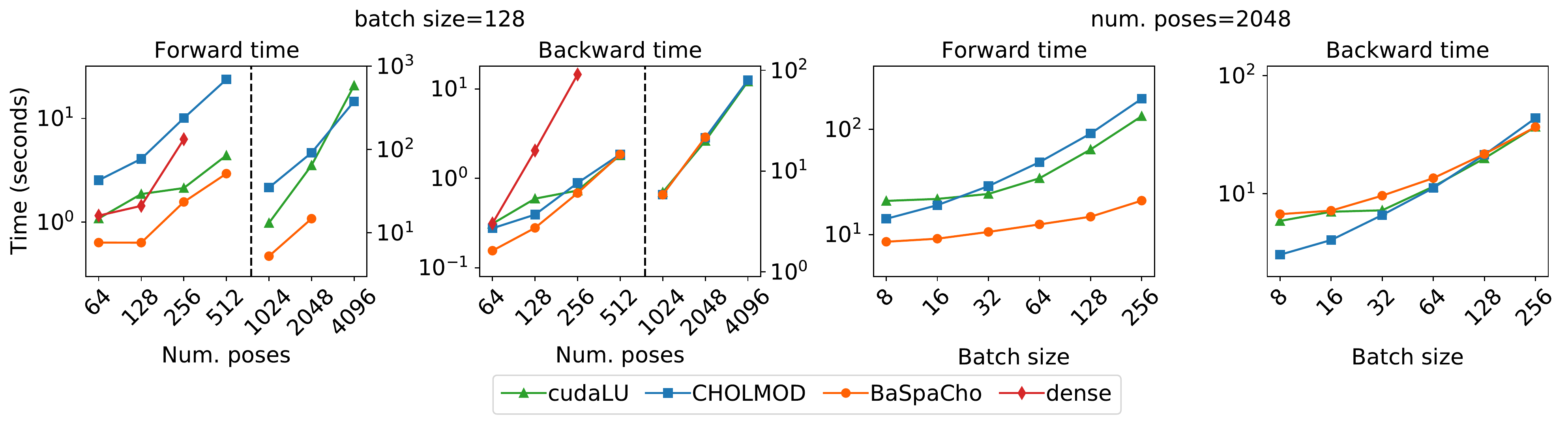}
    \vspace{-6mm}
    \caption{Forward/backward times of \theseus with sparse and dense solvers on different PGO problem scales.}
    \vspace{-1mm}
    \label{fig:profiling}
\end{figure}

\Cref{fig:profiling} shows the average time of a full forward and backward pass for a given batch size, taken by \theseus with different solvers (\cusparse, \cholmodp, \baspacho and \dense), for different problem scales (number of poses and batch size). The two left plots show time as a function of number of poses for a batch size of 128, while the two right plots show time as a function of batch size for 2048 poses. We find that \dense does not scale well with poses or batch size. For a batch size of 128, the largest problem that it can solve before running out of GPU memory has 256 poses (left two plots). With 2048 poses, \dense is unable to solve the problem regardless of batch size (right two plots). On the other hand with a batch size of 128, our solvers \baspacho scale to 2048 poses and \cusparse scale to 4096 poses. \cholmodp can solve problems even larger, since the linear system is solved on CPU and we have successfully tested up to 8192 poses and batch size 256 (see~\Cref{app:bench}), for a total of 22GBs of GPU usage for residuals and Jacobian blocks computation.

In addition to being more memory efficient, running times of our sparse solvers are also smaller for large enough number of poses/batch size, especially for the backward pass. Even though \dense's total time for forward+backward is comparable to \cusparse and faster than \cholmodp for smaller problems: \eg, 1.47s (\dense) vs. 1.32s (\cusparse) and 2.82s (\cholmodp) for batch size 128 and 64 poses, \dense is significantly slower or out of memory for larger problems. For the largest problem that \dense can solve (batch size 128 and 256 poses) its total time is already much slower than all others methods: 20.81s (dense) vs. 10.96s (\cholmodp), 2.86s (\cusparse), and 2.25s (\baspacho). Furthermore, \baspacho outperforms \dense for any problem scale and is up to one order of magnitude faster, including for smaller batch sizes and number of poses (see~\Cref{app:bench} for more results and details). For the largest problem that we consider (batch size 256 and 2048 poses), the total times for our sparse solvers are 170.28s for \cusparse, 239.07s for \cholmodp, and 57.67s for \baspacho.


\vspace{-3mm}
\subsection{Profiling \theseus as stand-alone \nls optimizer}\label{sec:ceres-comp}
\vspace{-2mm}

\dnls typically involves solving numerous optimization problems each epoch where a fast \nls optimizer is essential. We compare \theseus as a stand-alone \nls optimizer with the state-of-the-art \ceres \cite{agarwal2012ceres} library for solving a batch of PGO problems without any learning involved. We compare all solvers in terms of the total time required to perform 10 iterations on a set of 256 PGO problems. CPU/GPU configurations are same as before. For \texttt{CHOLMOD}, we also include a configuration that runs everything on CPU, including Jacobians and residual computation (labelled \texttt{CHOLMOD-allcpu}).

\begin{wrapfigure}{r}{0.58\textwidth}
    \vspace{-4mm}
    \includegraphics[width=0.58\textwidth]{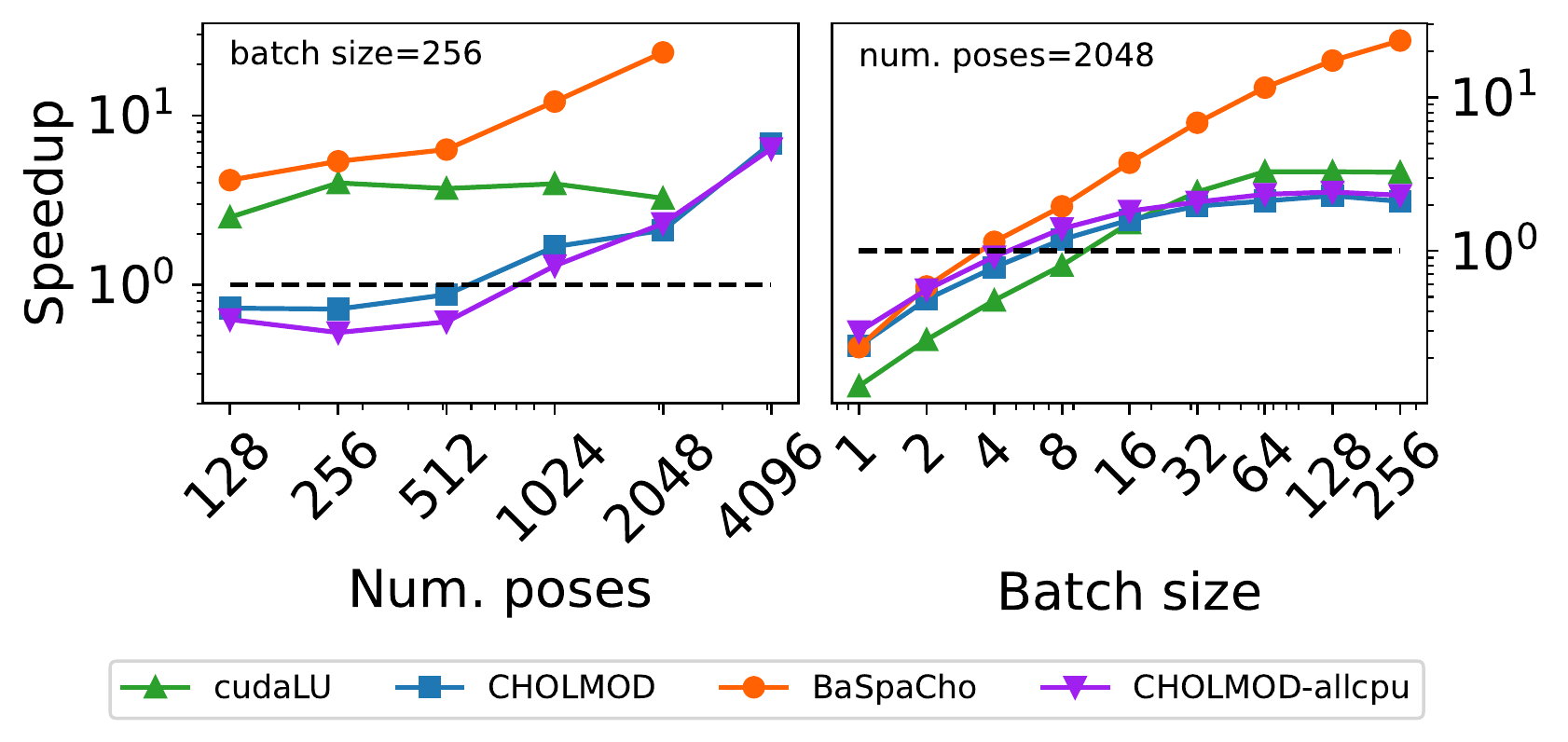}
    \vspace{-4mm}
    \caption{Speedup of \theseus (forward pass) over Ceres (black dashed) on different PGO problem scales.}
    \label{fig:ceres-comparison}
\end{wrapfigure}


\Cref{fig:ceres-comparison} shows speedup obtained by \theseus with batching, vectorization and sparse solvers, over \ceres as a function of increasing number of poses or batch size. We vary the number of poses for a fixed batch size of 256, and vary the batch size for a fixed number of poses of 2048.  Although \ceres is faster than all of our solvers when the number of poses and batch size are small (for instance, \ceres is 25x faster with 256 poses and 16 batch size, see~\Cref{app:bench}), as these increase \theseus shows significant speedup by being able to solve larger batches of problems in parallel. For the largest setup that all our solvers can scale to (2048 poses, 256 batch size), \baspacho is $\sim$23x faster than Ceres, and our other solvers are $\sim$4x faster. \cholmodp has a 6x speedup for its largest setting (4096 poses, 256 batch size).

Since typical use case of \theseus involves large batches and number of variables during end-to-end learning with \dnls, the speedups in this setting against a performant \nls solver highlights the significance of our efficiency-based design choices. See~\Cref{app:bench} for additional results of smaller fixed batch size and number of poses.


\vspace{-3mm}
\subsection{Backward modes analysis}\label{sec:backward-experiments}
\vspace{-2mm}

We explore the trade-offs between our different backward modes using the tactile state estimation application in~\Cref{sec:applications}. The learnable components here include a neural network, and thus closely follow the type of applications that motivate \theseus. We compare the following backward modes: derivative unrolling (\unroll), implicit differentiation (\implicit), truncated differentiation (\trunc), and direct loss minimization (\dlm); for \trunc we include results when truncating 5 and 10 steps. We compare all modes along 3 axis of performance: validation loss after 100 epochs (outer loop), run time during training, and peak GPU memory consumption of \texttt{TheseusLayer}. For these experiments we used Quadro GP100 GPUs with 16GB of memory. For time and memory we present separate results for forward and backward pass, and all numbers are averaged over 700 (7 batches for 100 epochs). Below we discuss our main findings from this analysis, and more results and details can be found in~\Cref{app:backward}.

\begin{figure}[t]
    \centering
    \includegraphics[width=\textwidth]{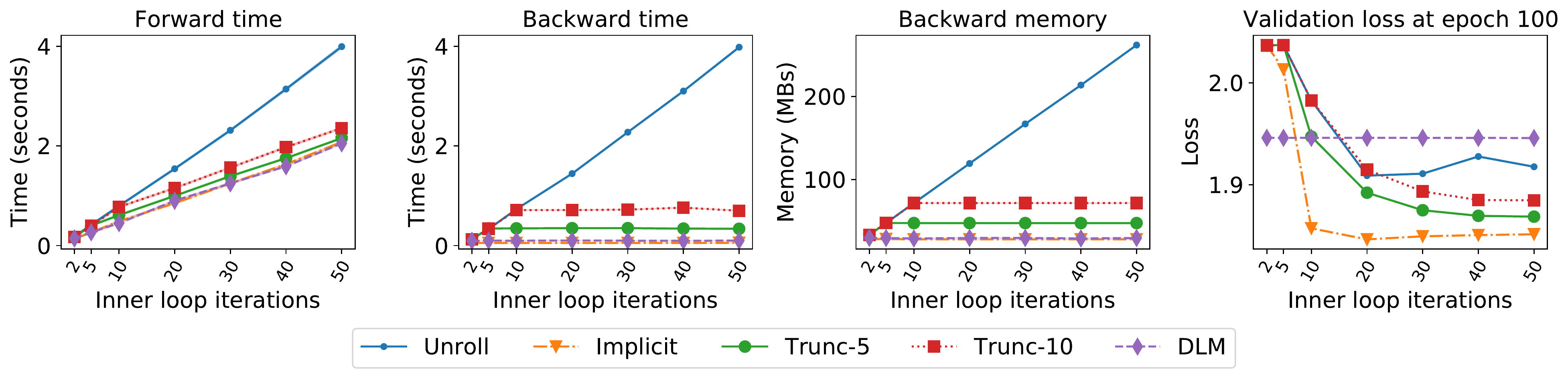}
    \vspace{-5mm}
    \caption{Time and memory consumption of different backward modes in tactile state estimation.}
    \label{fig:time-and-mem-tactile}
    \vspace{-1.5mm}
\end{figure}

\Cref{fig:time-and-mem-tactile} shows average run times for all backward modes as a function of the maximum number of iterations in the inner loop optimization. We observe that the time used in the forward pass (\Cref{fig:time-and-mem-tactile}, far left) increases roughly linearly for all modes, all having similar times except for \unroll, which is slower than other modes. On the other hand, we observe stark differences in the backward pass time (\Cref{fig:time-and-mem-tactile}, center left), where \unroll is the only method that has a linear dependence on the number of inner loop iterations. All other methods have a constant footprint for computing derivatives, independent of the number of inner loop iterations. As expected, increasing the number of iterations through which we backprop (5 or 10 for \trunc, all iterations for \unroll) increases the time necessary for a backward pass (\implicit = \dlm $<$ \truncN{5} $<$ \truncN{10} $<<$ \unroll).

Figure~\ref{fig:time-and-mem-tactile} (center right) shows the average peak memory consumption of the backward modes. In this case, the trends observed for the backward pass memory consumption is similar to the trends in time. In particular, \unroll's memory footprint increases linearly with the number of inner loop iterations, from $\sim34$MBs to $\sim262$MBs; for all other methods the memory consumption remains constant. The best memory profiles in this example is obtained with \implicit and \dlm backward modes, with $\sim28$MBs and $\sim29$MBs, respectively. These trends also hold for the forward pass memory consumption.

Figure~\ref{fig:time-and-mem-tactile} also shows the validation losses obtained with all backward modes (far right). The best validation loss, after 100 epochs of training, is obtained using \implicit, followed by \trunc variants. We notice that both variants of \trunc keep improving with increasing number of inner loop iterations, and that \unroll and \implicit achieve the best results with 20 iterations. One exception is \dlm, which doesn't improve much with the number of iterations, but is also the best method when only 2 inner loop iterations are performed.
%
%
As a point of caution, we stress that, unlike the timing and memory results, the relative training performance between different backward modes is likely to be application dependent, and is affected by hyperparameters such as the step size used for the inner loop optimizer (0.05 in this example), and the outer optimizer's learning rate. Our experiments suggest that implicit differentiation is a good default to use for differentiable optimization, considering its low time/memory footprint, and potential for better end-to-end performance with proper hyperparameter~tuning.

\vspace{-4mm}
\section{Discussion}
\vspace{-3mm}

\textbf{Summary.} \theseus provides nonlinear least squares as a differentiable layer and enables easily building and training end-to-end architectures for robotics and vision applications. We illustrate several example applications using the same application-agnostic interface and demonstrate significant improvements in performance with our efficiency-based design. Following how autodiff and GPU acceleration (among others) have led to the evolution of \pytorch in contrast to \texttt{NumPy}~\citep{harris2020array}, we can similarly view sparsity and implicit differentiation on top of autodiff and GPU acceleration as the key ingredients that power \theseus, in contrast to solvers like \ceres that typically only support sparsity. When solving a batch of large problems the forward pass of Theseus is up to 20x faster than \ceres.

\vspace{-0.5mm}

\textbf{Limitations.} \theseus currently has a few limitations. The nonlinear solvers we currently support apply constraints in a soft manner (\ie, using weighted costs). Hard constraints can be handled with methods like augmented Lagrangian or sequential quadratic programs~\citep{Sodhi2020, Martins2011}, and differentiating through them are active research topics. The current implementation of LM does not support damping to be learnable. Some limitations and trade-offs with the sparse linear solvers are discussed in~\cref{sec:sparse-solver}, and with backward modes are discussed in~\cref{app:backward}. Online learning applications may require frequently editing the objective and depending on the problem size there may be a nontrivial overhead that is not currently optimized as we explored only non-incremental settings in this work. Additional performance gains can be extracted by moving some of our Python implementation to C++ but we prioritized flexibility in evolving the API in the short-term.
We do not yet support distributed training beyond what \pytorch natively supports. We will explore these features and optimizations in the future as the library continues to evolve.



\ifARXIV
\vspace{-4mm}
\section*{Acknowledgments}
\vspace{-3mm}
The authors would like to thank Dhruv Batra, Olivier Delalleau, Jessica Hodgins, and Mary Williamson for guidance and support on the project, Dhruv Batra and Sal Candido for feedback on early drafts of the paper, Franziska Meier for help with the differentiable robot model library, Paul‑Edouard Sarlin for helpful discussion on the homography example, Horace He, Richard Zou and Samantha Andow for help with \functorch~\citep{functorch2021} library, Terran Washington, Gopika Jhala and Chantal Mora for designing the Theseus logo, and Oliver Libaw, Christine Gibson, Orialis Valentin, Alyssa Newcomb and Eric Kaplan for help with the blog post. The authors also thank community members for contributions to the open source code. Work by PS and JO was done while at Meta AI.

\fi

\ifCR
\vspace{-4mm}
\begin{ack}
\vspace{-3mm}
The authors would like to thank Dhruv Batra, Olivier Delalleau, Jessica Hodgins, and Mary Williamson for guidance and support on the project, Dhruv Batra and Sal Candido for feedback on early drafts of the paper, Franziska Meier for help with the differentiable robot model library, Paul‑Edouard Sarlin for helpful discussion on the homography example, Horace He, Richard Zou and Samantha Andow for help with \functorch~\citep{functorch2021} library, Terran Washington, Gopika Jhala and Chantal Mora for designing the Theseus logo, and Oliver Libaw, Christine Gibson, Orialis Valentin, Alyssa Newcomb and Eric Kaplan for help with the blog post. The authors also thank community members for contributions to the open source code. Work by PS and JO was done while at Meta AI.

\end{ack}
\fi




\ifARXIV{\ifAPP{\appendix

\section*{Appendix}
\vspace{-2.5mm}

\section{Contributions}\label{sec:contrib}
\vspace{-3mm}

The contributions of the authors are as follows.\\
\textbf{Luis Pineda} led the engineering of the project, developed and implemented the core API, differentiable nonlinear solvers, motion planning example and tutorials, standard and autodiff cost functions, and backward mode experiments, coordinated with sub-teams to help design, implement, integrate and review of all aspects of the code and evaluations, wrote the paper.\\
\textbf{Taosha Fan} developed and implemented differentiable Lie groups, automatic vectorization, \functorch support, pose graph optimization example, performance evaluations and benchmarking, helped with API design, bug fixes, did code reviews, wrote the paper.\\
\textbf{Maurizio Monge} developed and implemented all sparse linear solvers, \baspacho solver and bundle adjustment example, batching, differentiation and custom C++/CUDA backends for sparse solvers, did code reviews, wrote sparse solver and bundle adjustment sections in the paper.\\
\textbf{Shobha Venkataraman} helped develop and implement the motion planning example, helped with API design, wrote several tutorial notebooks.\\
\textbf{Paloma Sodhi} developed and implemented the tactile state estimation example, energy based learning and covariance sampling example, helped design learning API.\\
\textbf{Ricky T. Q. Chen} developed and implemented the direct loss minimization backward mode, wrote its section in the paper.\\
\textbf{Joseph Ortiz} developed and implemented the homography example, helped write its section in the paper, did some bug fixes, helped with the implicit backward mode theory.\\
\textbf{Daniel DeTone} prototyped the homography example, helped develop its final version, wrote its section in the paper.\\
\textbf{Austin Wang} implemented differentiable forward kinematics support, wrote its section in the paper.\\
\textbf{Stuart Anderson} advised on the project, managed and supported research engineers, helped edit the paper.\\
\textbf{Jing Dong} advised on the project, helped design the core API and performance evaluations, did code reviews, helped edit the paper.\\
\textbf{Brandon Amos} advised on the project, developed and implemented implicit and truncated backward modes for nonlinear solvers, helped implement implicit backwards for linear solvers, helped design and analyze backward mode experiments, did code reviews, wrote the backward modes sections and helped edit the paper.\\
\textbf{Mustafa Mukadam} led the project, set the vision and research direction, created and steered the team, provided guidance on all aspects including feature prioritization, API design and implementation, formulated evaluations and experiments, did code reviews, wrote the paper.

\vspace{-3mm}
\section{Simple example description}\label{sec:simple-example}
\vspace{-3mm}

In this section, we describe the example in~\Cref{lst:basic_differentiable} in more detail. The example considers fitting the curve $y = ve^x$ to a dataset of $N$ observations $(x,y)\sim\mathcal{D}$. A standard way to solve this is to minimize the least squares objective~\Cref{eq:nls} with residuals $r_i(\hat{v}) \defeq y^{(i)} - \hat{v}e^{x^{(i)}}$, for $i=1,...,N$, and where $\theta \defeq \hat{v}$. We can model this in \theseus with a single \cf that computes the $N$-dimensional vector $R(\hat{v})$ of all residuals as a function of a single optimization \texttt{Variable} $\hat{v}$ and two auxiliary variables, $x$ and $y$. 

The code implementing this problem starts by creating uniquely named \texttt{Variable} containers in lines 3-5. We then create an objective with the cost function (lines 11-15). We use a \cf of type \texttt{AutoDiffCostFunction} (line 12), which relies on \texttt{torch.autograd} and vectorization via \functorch~\citep{functorch2021} to automatically compute the residual Jacobians used by the inner optimizer (see \cref{app:adc}). \texttt{AutoDiffCostFunction} requires providing an error function that receives optimization variables and auxiliary variables (defined in lines 7-9), and returns \texttt{torch} tensors computing the (unweighted) residual. Although not required by this problem, we also illustrate how to add a cost weight to the residuals by including a \texttt{ScaleCostWeight}, which simply scales all residuals in this cost function by a scalar (1.0 in this case). Finally, we encapsulate the objective and a Gauss-Newton optimizer into a differentiable \thlayer in line 16. 

To illustrate how to differentiate through this layer, we perturb the $x$ values in the dataset so that it becomes impossible to recover the correct value of $v$ from \nls optimization alone, and then define $\phi \defeq x$ and \mbox{$L(\theta^*(\phi)) \defeq ( \hat{v}^*(x) - v)^2$}. Thus, the outer loop optimization corrects the $x$ tensor so that the solution of the inner loop matches the true value of $v$. 

In the code, the outer parameter is defined in line 18, with initial value for $\phi$ set to a perturbed function of the true $x$, while the outer loss is defined in line 24. In lines 21-23 we solve the \nls problem, by calling \texttt{forward()} with the current value of $\phi$ as the value to set for auxiliary variable named ``x'', and an initial value $v = 1$ for the optimization variable named ``v'' (with a required batch dimension in the input); we also set the backward mode to \implicit. The optimum found can be recovered as a tensor by getting key ``v'' of the dictionary returned by \texttt{forward}, which we then use to compute the outer loss in line 24. Finally, outer loop optimization is done via \texttt{torch}'s well-known autograd engine, in lines 25-26, using the Adam optimizer~\citep{kingma2014adam} for $\phi$, defined in line 19.

\vspace{-3mm}
\section{\adc autograd modes}\label{app:adc}
\vspace{-3mm}

In this section, we evaluate the effect of the three different autograd modes we provide for automatically computing the jacobians for \adc:
\vspace{-0.5mm}
\begin{itemize}[leftmargin=5mm,topsep=0pt,itemsep=2.5pt]
    \item \textbf{dense} uses \texttt{torch.autograd.functional.jacobian}, which computes a dense jacobian that includes cross-batch derivatives; \ie, the derivative of $j$-th batch output with respect to variables in the $i$-th batch input. Since we only need per-sample gradients, we slice the result of this operation.
    \item \textbf{\loopbatch} is also based on \texttt{torch.autograd.functional.jacobian}, but we manually loop over the batch before each call so that we obtain only per-sample gradients. 
    \item \textbf{vmap} mode uses \texttt{functorch.vmap}~\citep{functorch2021} to compute per-sample gradients in a vectorized manner. 
\end{itemize}

For evaluation, we use the homography example described in~\cref{sec:homography}, which uses \adc and can have significant memory requirements when computing the jacobians. \cref{fig:autograd_modes} illustrates the advantages of using \textbf{vmap} over the other two modes, both in terms of compute time and memory. The \textbf{\loopbatch} mode has similar memory requirements to \textbf{vmap}, but the compute time is significantly slower than the other two methods. Finally, \textbf{dense} mode has substantially more memory requirements than the two other methods (up to an order of magnitude higher), and runs out of memory for the largest batch size 256 used in this experiment. For a batch size of 128, \textbf{vmap} results in a speedup of 22x over the next best method ($\sim15$ms vs. $\sim337$ms for \textbf{dense}), and almost 8x less memory ($\sim1.2$GBs vs. $\sim9.6$GBs).

\begin{figure}[!t]
    \centering 
    \includegraphics[width=0.6\textwidth]{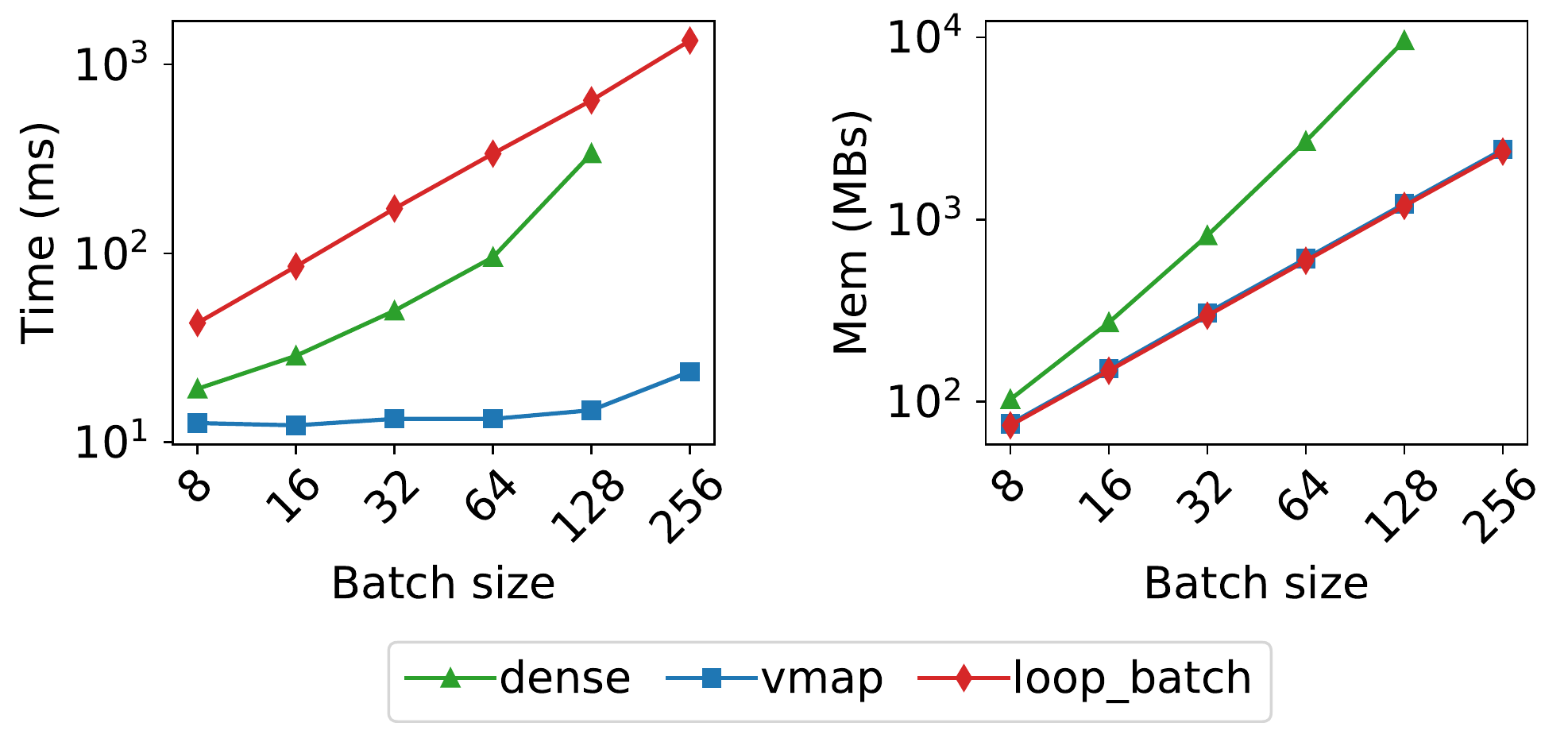}
    \vspace{-1.5mm}
    \caption{\adc time (left) and memory (right) consumption in the homography estimation example for different autograd modes and batch sizes.}
    \label{fig:autograd_modes}
\end{figure}

\vspace{-3mm}
\section{Differentiable Lie group details}\label{app:lie}
\vspace{-3mm}

While differentiation on the Euclidean space is straightforward, it remains challenging to do so on the non-Euclidean manifolds. In this section, we provide details about how to compute the derivatives on the tangent space of Lie groups using the projection operator~\citep{absil2009optimization}. The implementation of the projection operator is essential for automatic differentiation on the tangent space of Lie groups.

Suppose $F(g)$ is a function of $g\in G$ where $G$ is a matrix Lie group and $\tau(\xi)$ a retraction map of $G$. For notational simplicity, let $\nabla_g F(g)$ denote the Euclidean gradient of $F(g)$ and $T_e G$ the Lie algebra of $G$. Following~\citep{chirikjian2011stochastic}, the gradient on the tangent space of Lie group is a linear operator $\D_g F(g)$ such that
\begin{equation}\label{eq:grad1}
    \D_g F(g)\cdot \xi = \left.\frac{\partial}{\partial s}\right|_{s=0} F\big(g\tau(s\cdot \xi)\big)
\end{equation}
holds for any Lie algebra elements $\xi\in T_e G$. As a result of the chain rule, the right-hand side of the equation above is equivalent to 
\begin{equation}\label{eq:grad2}
    \left.\frac{\partial}{\partial s}\right|_{s=0} F\big(g\tau(s\cdot \xi)\big)=\nabla_g F(g)\cdot \left.\frac{\partial}{\partial s}\right|_{s=0} g\tau(s\cdot\xi) =\nabla_g F(g)\cdot g\xi
\end{equation}
where the last equality results from properties of the retraction map. Then, we conclude from \Cref{eq:grad1,eq:grad2} that
\begin{equation}
    \D_g F(g)\cdot \xi = \nabla_g F(g)\cdot g\xi.
\end{equation}
Therefore, there exists a projection operator $\mathrm{proj}_g(\cdot)$ such that
\begin{equation}
\D_g F(g) = \mathrm{proj}_g \big(\nabla_g F(g)\big)
\end{equation}
for any gradients on the tangent space of Lie group and their corresponding Euclidean gradients~\citep{absil2009optimization}. Furthermore, note that the projection operator $\mathrm{proj}_g(\cdot)$ is a linear operator depending on $g\in G$ and can be computed in closed form.

\vspace{-2.5mm}
\section{Example application details}\label{app:examples}
\vspace{-2.5mm}

\subsection{Pose graph optimization}\label{app:pgo}
\vspace{-2.5mm}

Pose graph optimization (PGO)~\citep{rosen2019SESync,fan2020cpl,fan2021majorization} is the problem of recovering unknown poses of SE(2) and SE(3) from the noisy relative pose measurements. Pose graph optimization has extensive applications in robotics~\citep{grisetti2011g2o}, computer vision~\citep{martinec2007robust}, computational biology~\citep{singer2011three}, sensor networks~\citep{cucuringu2012sensor}, etc. In pose graph optimization, we represent unknown poses as vertices and relative pose measurements as edges. Then, it is possible to compute the relative pose errors for each pair of neighboring vertices such that a nonlinear least-squares optimization problem can be formulated for pose estimation.  A more detailed introduction to pose graph optimization can be found in \cite{grisetti2011g2o,rosen2019SESync,fan2020cpl,fan2021majorization}.

\theseus includes a differentiable and coordinate-independent version of the relative pose errors with which it is straightforward to solve pose graph optimization. We evaluated \theseus on the simulated Cube dataset and a number of benchmark datasets for pose graph optimization~\citep{grisetti2011g2o,rosen2019SESync}. The Cube dataset simulates the 3D odometry of a robot with varying numbers of poses, loop closure probabilities, and loop closure outlier ratios, which is used to profile the time and space complexities of the forward and backward passes in \theseus. Furthermore, the benchmark datasets indicate that \theseus is capable of solving large-scale differentiable nonlinear optimization problems with comparable accuracy and efficiency to existing state-of-the-art solver like \ceres~\citep{agarwal2012ceres}. 

\theseus and \ceres attain the same objective values for all the evaluated benchmark datasets \citep{grisetti2011g2o,rosen2019SESync} using the chordal initialization~\citep{carlone2015init}. The inital/final objective values and qualitative results for some benchmark datasets are shown in \cref{tab:pgo_obj} and \Cref{fig:pgo}, respectively.

\begin{table}
	\centering
	\setlength{\tabcolsep}{0.285em}
	\begin{tabular}{c| c c c c c c}
		\toprule
		\multirow{2}{*}{}
		& \multicolumn{6}{c}{Objective Value} \\
		\cmidrule{2-7}
		& Sphere  & Torus & Cubicle & Rim & Grid & Garage   \\
		\midrule
		 Initial &  $8.437\!\times\!10^2$  & $1.234\!\times\!10^4$ & $1.622\!\times\!10^6$ & $1.924\!\times\!10^7$ &$4.365\!\times\!10^4$ & $7.108\!\times\!10^{-1}$    \\
		Final      &  $6.805\!\times\!10^2$  & $1.212\!\times\!10^4$ & $1.455\!\times\!10^3$ & $4.157\!\times\!10^4$ &$4.218\!\times\!10^4$ & $6.342\!\times\!10^{-1}$   \\
		\bottomrule
	\end{tabular}
	\vspace{1em}
	\caption{Initial and final objective values of \theseus on 3D benchmark datasets with PGO example.}
	\vspace{-5mm}
	\label{tab:pgo_obj}
\end{table}

\begin{figure}
\begin{tabular}{cccc}
	\subfloat[][Sphere]{\includegraphics[trim =0mm 0mm 0mm 0mm,width=0.14\textwidth]{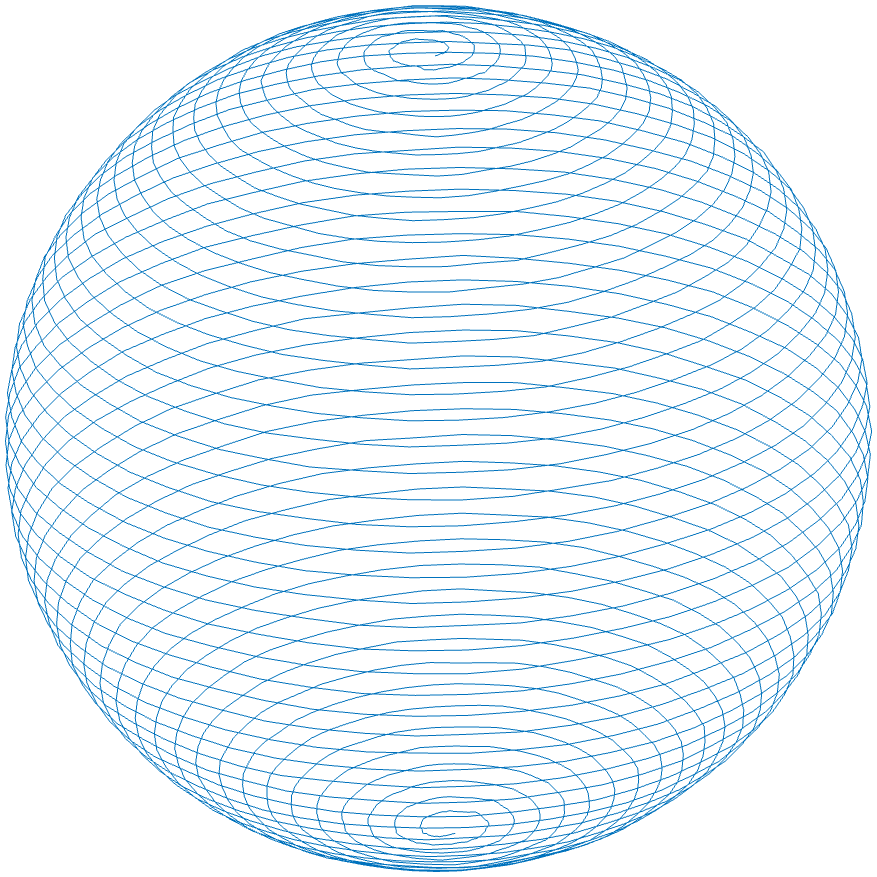}} &
	\subfloat[][Cubicle]{\includegraphics[trim =35mm 5mm 35mm 0mm,width=0.22\textwidth]{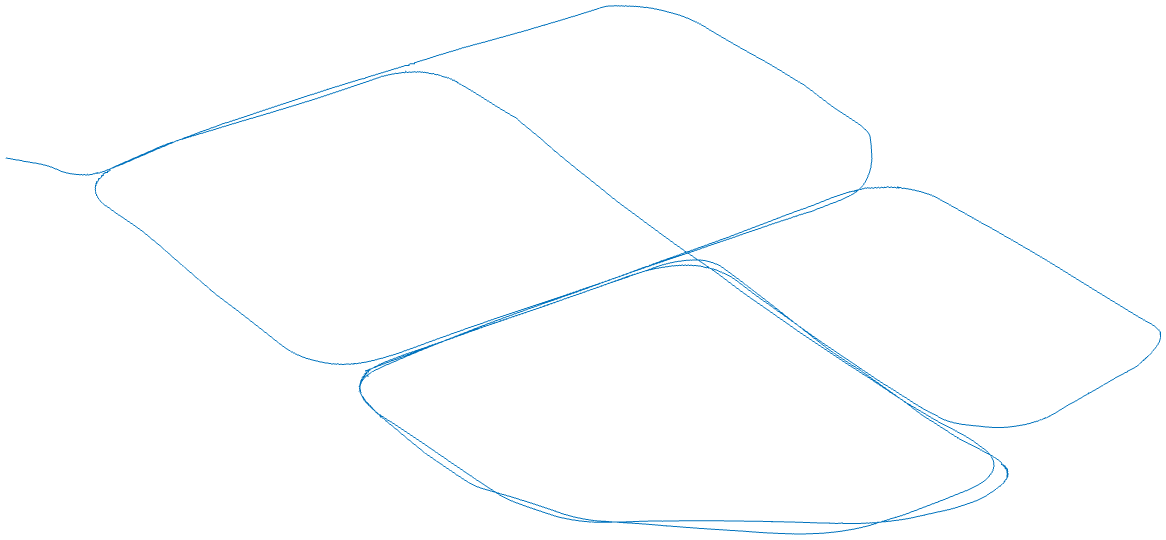}} &
	\subfloat[][Rim]{\includegraphics[trim =0mm 0mm 0mm 0mm,width=0.24\textwidth]{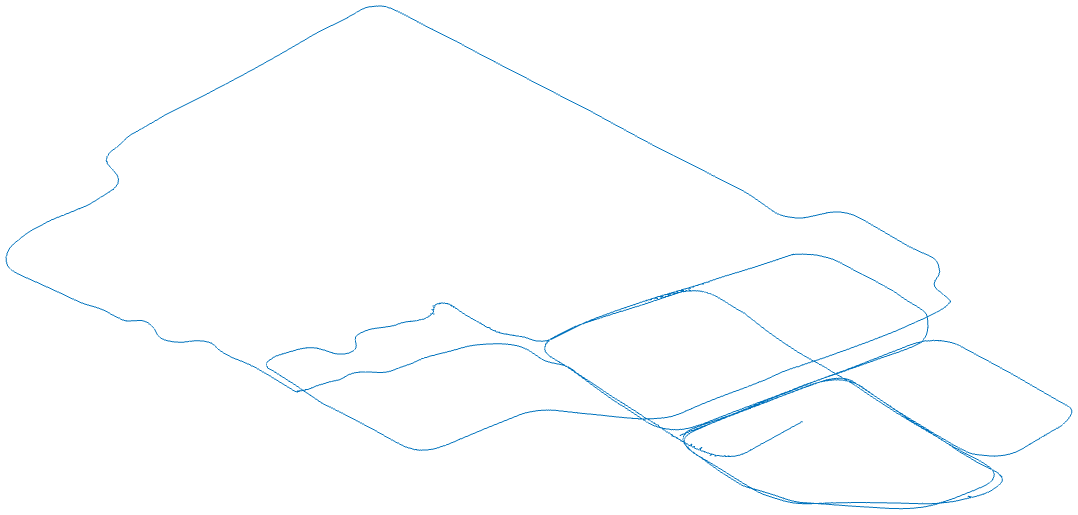}}&
	\subfloat[][Garage]{\includegraphics[trim =0mm 0mm 0mm 20mm,width=0.25\textwidth]{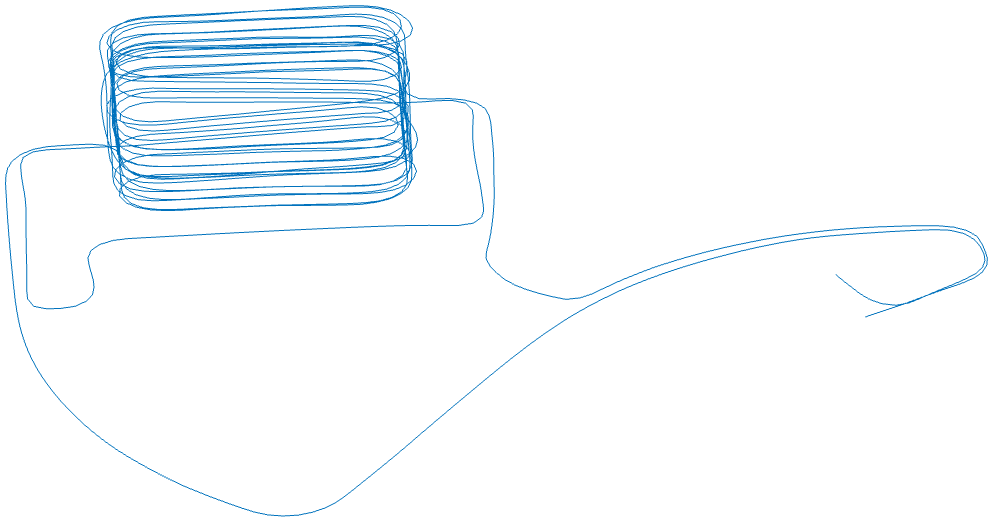}}
\end{tabular}
\vspace{6mm}
\caption{Qualitative results of \theseus on 3D benchmark datasets with PGO example.}
\label{fig:pgo}
\end{figure}

\vspace{-2.5mm}
\subsection{Tactile state estimation}
\label{app:tactile-problem}
\vspace{-2.5mm}

Recent work~\citep{sodhi2021learning} explored the use of \nls optimization with learned tactile sensor observations for tactile pose estimation. The goal is to incrementally estimate sequences of object poses that are moved by a robotic hand equipped with a DIGIT tactile tensor~\citep{lambeta2020digit}. The key insight of \citet{sodhi2021learning} is to use learning to transform high dimensional tactile observations into relative poses between measurement pairs. Once relative poses are available, the sensor data can be integrated into an optimization problem that solves for object poses. The objective includes four types of cost functions. One penalizes inconsistencies with the measurement coming from the learned observation model. A second one encourages the predicted poses to be consistent with a quasi-static physics model~\citep{ZhouBM17}. A third type adds geometric constraints by penalizing intersections between the end effector and the object using a signed distance field. Finally, a fourth cost function incorporates pose priors from a camera. 

In \theseus, we implement an offline and differentiable version of the tactile state estimation problem above, using a dataset of 63 trajectories of length 25 with known ground truth poses provided by the authors of~\citep{sodhi2021learning}; we used 56 of these as training set and the other 7 as a test set. Optimization variables are object and end effector poses (modeled as SE(2) groups) for each point in the trajectory, and the outer loss objective is the difference between the optimized object poses and the ground truth in the dataset.  The learnable component corresponds to the relative pose model, using a pre-trained encoder, and finetuning the final layer via end-to-end learning through the inner loop optimization. This approach is similar to how the tactile measurements model was trained in~\citep{sodhi2022leo}, with the two main differences being that we do not use an energy-based formulation and instead directly differentiate through the inner optimizer, and we also do not consider an incremental setting. \Cref{fig:tactile-and-motion-examples} (left) shows an example of estimated trajectories before and after learning.

\vspace{-2.5mm}
\subsection{Bundle adjustment}
\vspace{-2.5mm}

Bundle adjustment is the problem of optimizing a 3D reconstruction formed by a set of camera images and a set of landmarks observed and matched across the images. In every camera image a 2D coordinate is identified for the position of all observed landmarks, and the problem is initialized with an estimate of the positions of the landmarks and the camera poses. We call \emph{reprojection error} the image-offset between where the landmark was detected on the image, and the reprojection of the landmark according to the current parameter estimation. The optimization problem consists of simultaneously tweaking the cameras poses and landmark positions, while minimizing the square-sum of all the reprojection errors; see \cite{triggs1999bundle, agarwal2010bundle} for in depth exploration on bundle adjustment and its state-of-the-art.

We provide a bundle adjustment application example in \theseus, adopting the same data format of \cite{agarwal2010bundle}, with functions to generate synthetic dataset, as well as load/save open source datasets. To test bundle adjustment in a differentiable optimization setting, we add soft-kernels to the reprojection errors and setup as outer loop parameter the radius of the soft-kernel, which represents the confidence radius for reprojection errors with respect to possible outlier observations. We use as outer loss the average frame pose error from a ground truth value, such that the outer loop's task is to set the radius to a value that will make the bundle adjustment problem set the ideal soft loss radius value.

\begin{figure}
    \centering
    \begin{subfigure}[b]{0.24\textwidth}
        \includegraphics[trim=20mm 0mm 10mm 0mm,clip,width=\textwidth]{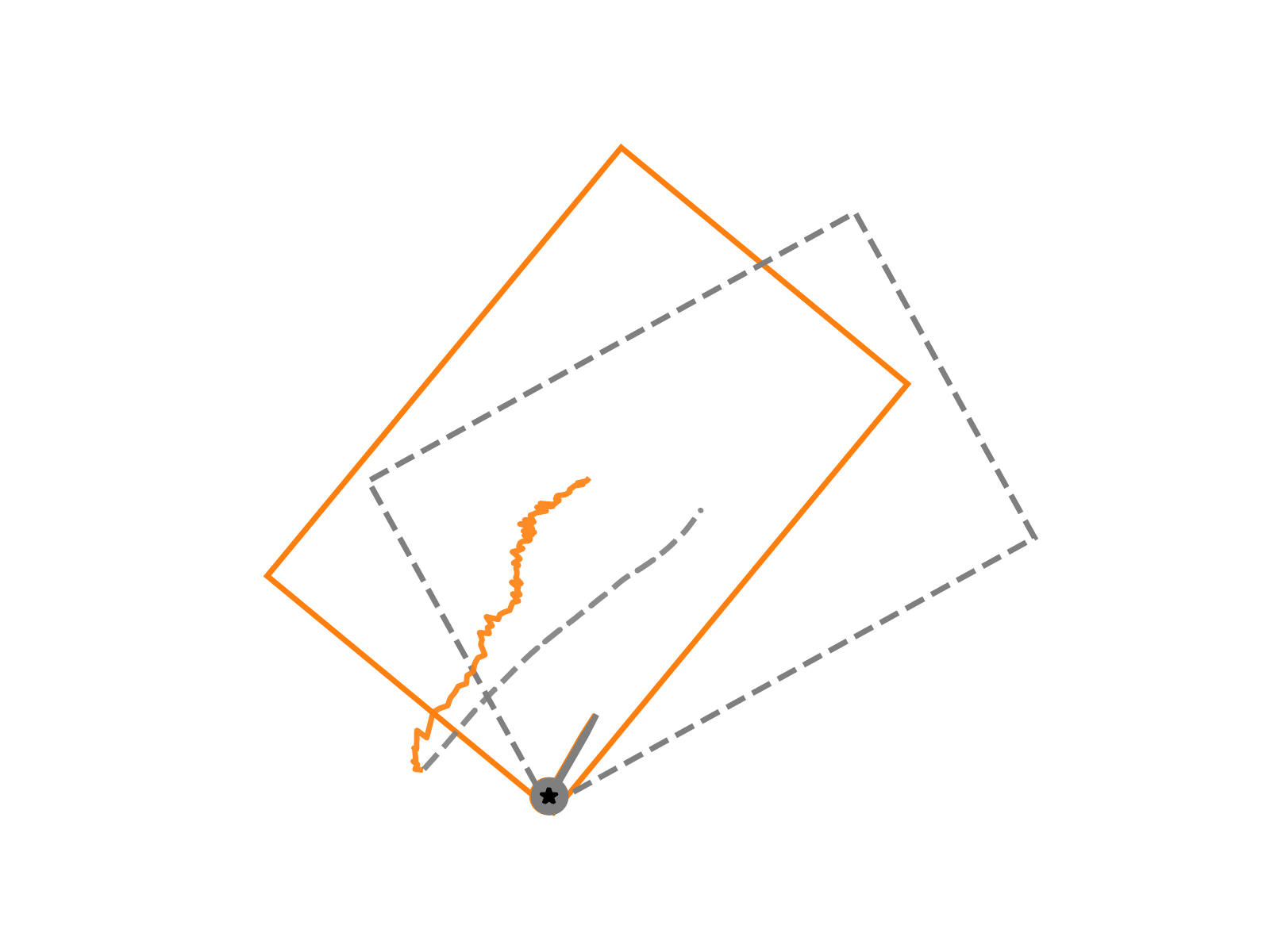}
    \end{subfigure}
    \kern-1.5em
    \begin{subfigure}[b]{0.24\textwidth}
        \includegraphics[trim=20mm 0mm 10mm 0mm,clip,width=\textwidth]{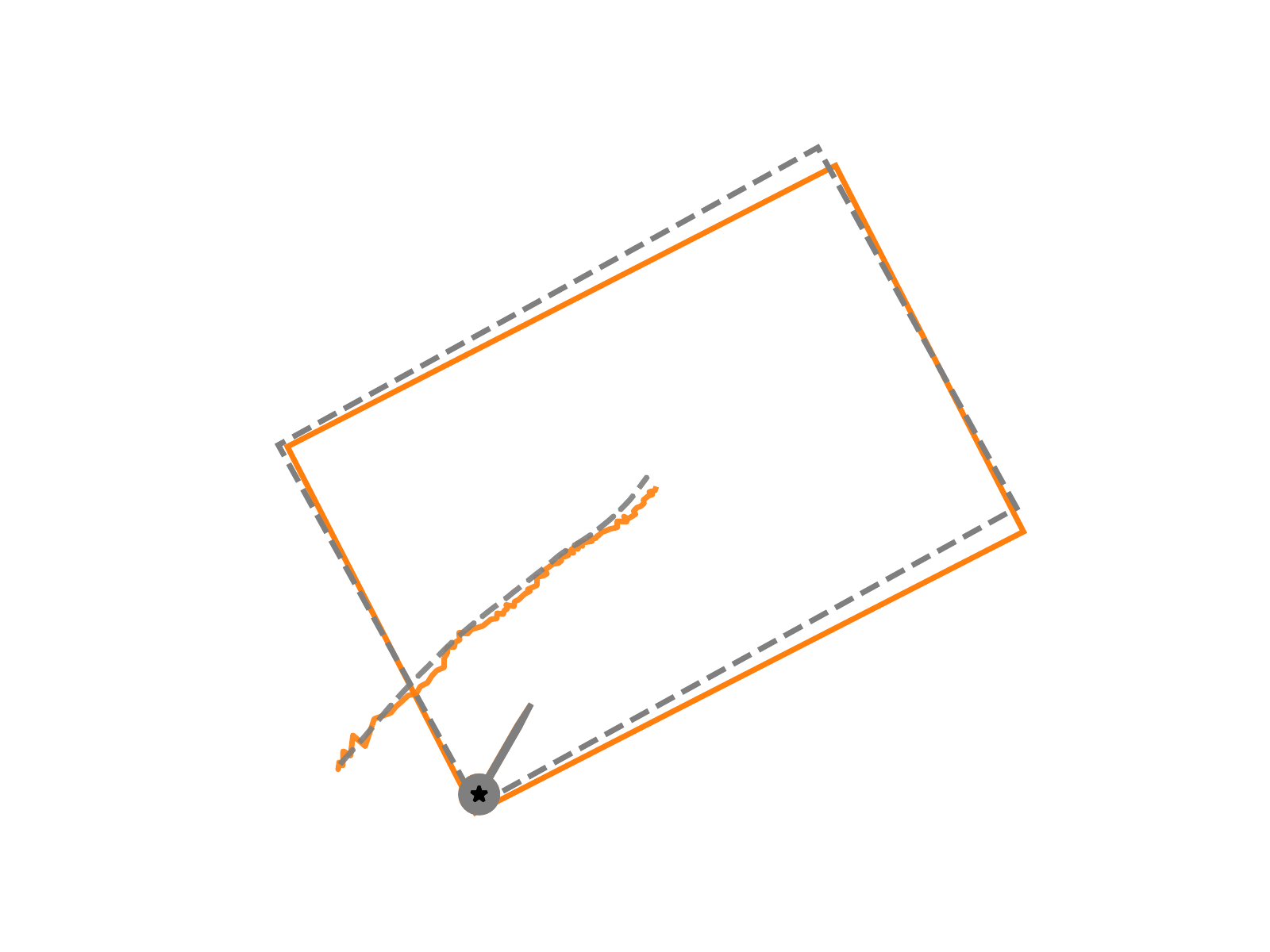}
    \end{subfigure}
    \hspace{10mm}
    \begin{subfigure}[b]{0.2\textwidth}
        \includegraphics[width=\textwidth]{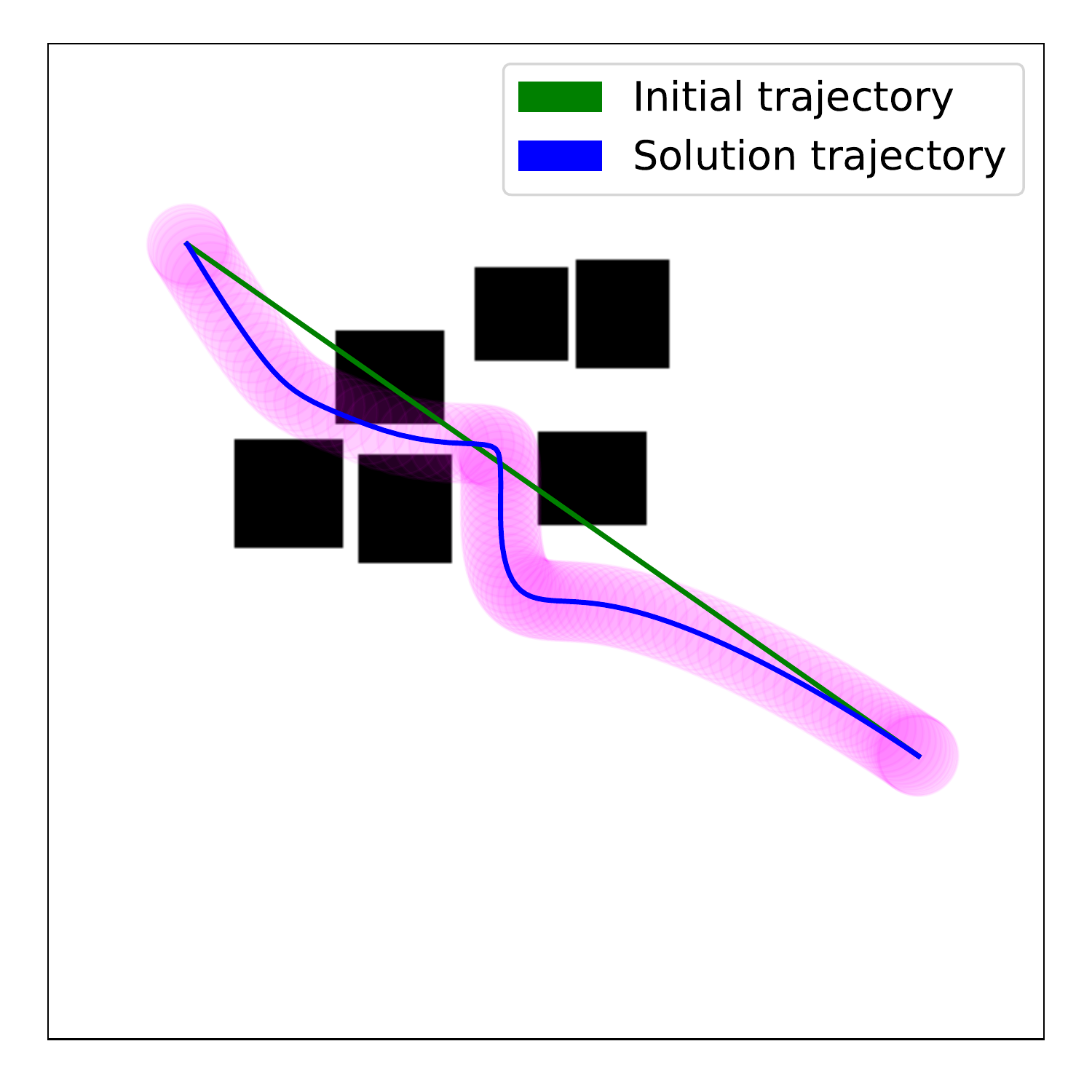}
    \end{subfigure}
    \kern-0.5em
    \begin{subfigure}[b]{0.2\textwidth}
        \includegraphics[width=\textwidth]{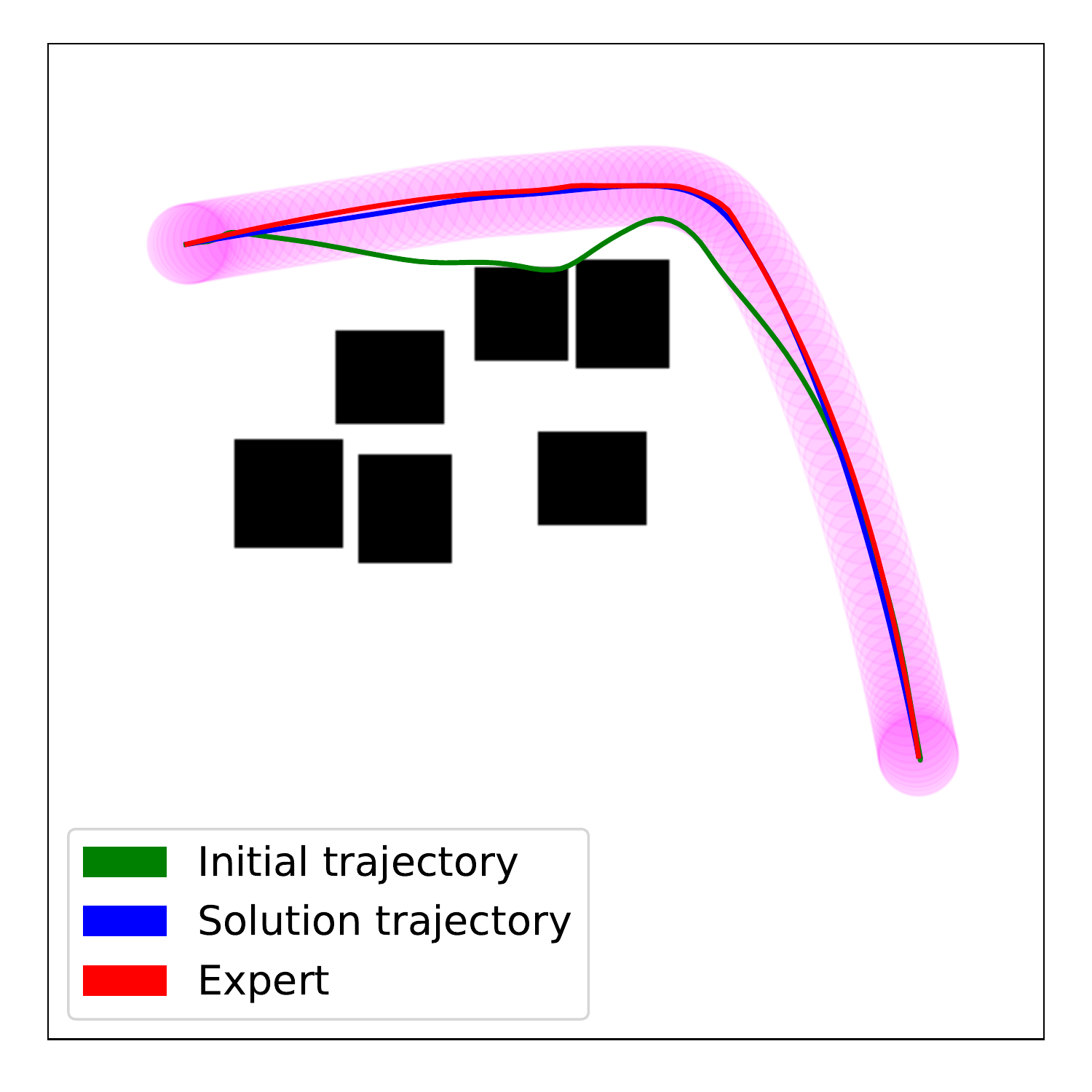}
    \end{subfigure}
    \caption{Examples of differentiable tactile state estimation and differentiable motion planning with \theseus. \textbf{Left:} Pose estimates before and after learning in the tactile state estimation example. Grey color indicates ground truth, and orange the estimate. The plot shows the trajectory as a curve, and the rectangle indicates the last object pose in the trajectory. \textbf{Right:} Trajectories generated by the planner before and after learning. The learned model generates initial trajectories for the optimizer to match an expert with only a few optimization~steps.}
    \label{fig:tactile-and-motion-examples}
    \vspace{-1mm}
\end{figure}

\vspace{-2.5mm}
\subsection{Motion planning}
\vspace{-2.5mm}

\nls optimization can also be used for motion planning in robotics~\citep{mukadam2018continuous}, where the objective variables are robot poses and velocities on a set of discrete time steps. Cost functions include terms representing smoothness constraints modeling forward kinematics, collision avoidance penalties, and boundary conditions on start and goal states. An end-to-end differentiable version of this formulation was proposed by~\citet{bhardwaj2020differentiable}, where a neural model predicts state-dependent cost weights for each step in the path, and the outer loss encourages the inner loop optimization to produce paths matching an expert in a dataset of trajectories. As part of \theseus, we include differentiable versions of cost functions like smoothness and collision in~\cite{mukadam2018continuous}, and an example of how to setup end-to-end differentiation for optimization variable initialization. That is, the model proposes initial trajectories for the optimizer, and the outer loss is set to a weighted sum of two terms, one computing closeness to the expert trajectory, and another equal to the inner loop's objective after only 2 iterations. This loss encourages the model to produce high-quality ``proposals'' that converge to good quality solutions quickly; an example of before/after training initial trajectories is illustrated in~\Cref{fig:tactile-and-motion-examples}. 

\vspace{-2.5mm}
\subsection{Homography estimation}
\label{sec:homography}
\vspace{-2.5mm}

A homography, also known as a projective transformation, is a linear mapping between a 2D point in one image $x$ to a point in another image $x'$, defined by a $3 \times 3$ matrix $H$, written as $x \sim H x'$, where $\sim$ defines the equivalence up to scale. In addition to representing linear transformations across 2D images, the homography is also a valid approximation of the motion of 2D points observed from camera images in 3D scenes in certain scenarios such as (1) rotation-only motion between cameras (2) when the scene is planar and (3) when the scene structure is far from the camera.

\begin{figure}
    \centering 
    \includegraphics[width=0.8\textwidth]{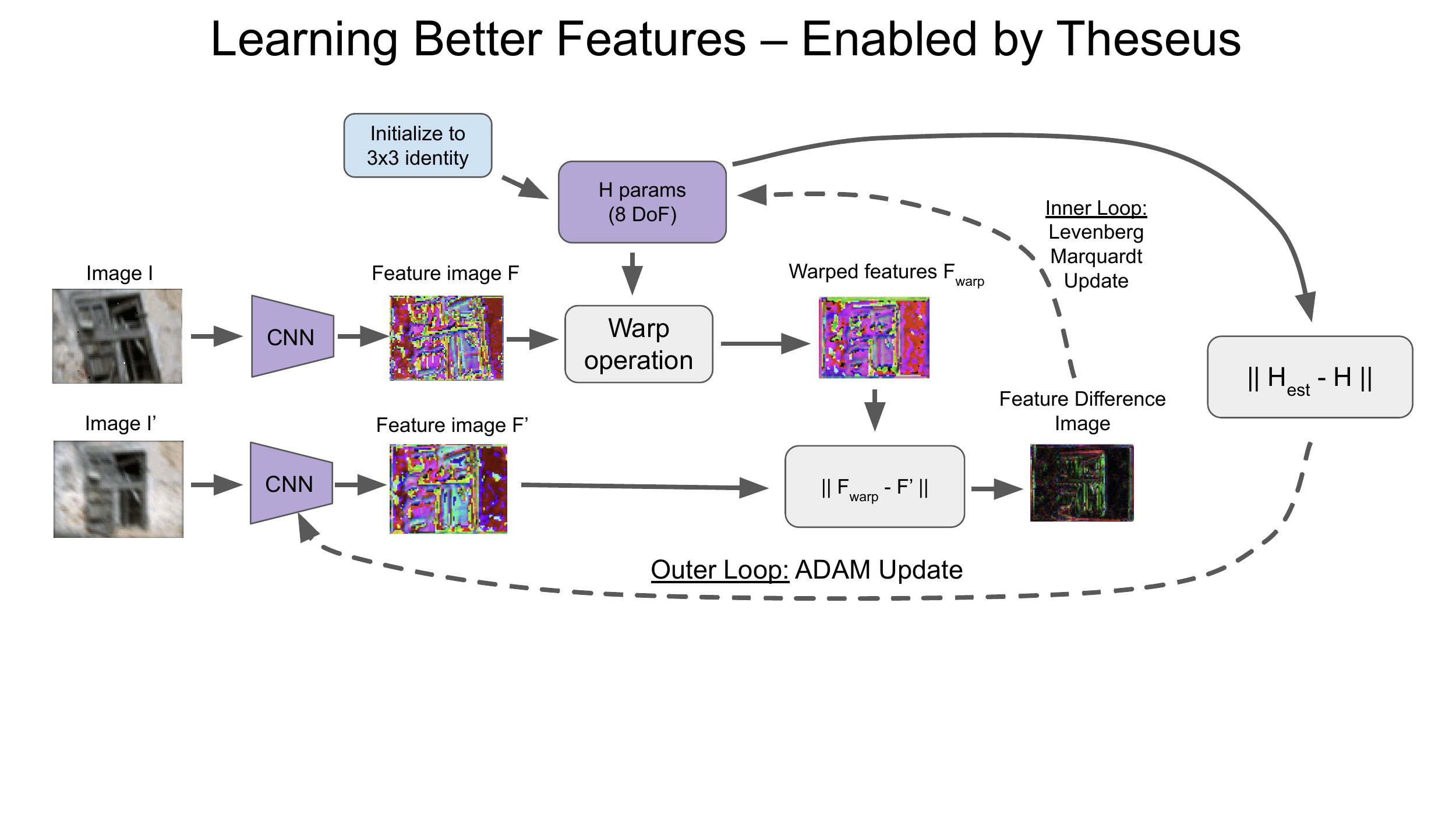}
    \caption{\textbf{Learning Robust Image Features for Homography Estimation with \theseus.} An inner loop optimization problem aligns two images via a feature-metric Levenberg-Marquardt optimization using features computed from a CNN. The outer loop uses Adam to update the weights the of the CNN that best minimize the final homography error.}
    \label{fig:homography}
    \vspace{-1mm}
\end{figure}

One approach to solving for the parameters of the homography is through iterative optimization via dense alignment of RGB image pixels in the image through fast second order optimization methods, as is done in Lucas-Kanade optical flow algorithm \citep{lucas1981, baker04}. This approach is also known as photo-metric alignment. Though this technique performs well in many scenarios, photo-metric alignment struggles when the lighting in the scene changes significantly, because it assumes that the brightness of a pixel is constant across different views. Feature-metric optimization is an extension to photo-metric optimization that works by first passing the image $I$ through a feature extractor function $f(\cdot)$, such as a convolutional neural network parameterized by weights $w$,  $F = f(I; w)$. This function generates a feature map $F \in \mathbb{R}^{C \times H \times W}$, where $C$ is a high dimensional channel number like $32$, and $H$ and $W$ represent the image height and width respectively. In feature-metric alignment, the alignment is done at the feature map level, rather than the RGB image level.

One important question when designing a feature-metric optimization algorithm, is how to obtain the weights $w$ that define the feature extractor. One approach used in works such as \citep{czarnowski2017} uses an off-the-shelf CNN which has been trained for image classification. One benefit of using \theseus for such as task is that the learning problem can be written without deriving analytical gradients, making it much easier to rapidly prototype and explore various formulations. In our example, we demonstrate a use-case of \theseus by performing end-to-end training of a two-layer CNN using gradients obtained through the homography optimization. A high level diagram of this learning problem is presented in~\cref{fig:homography}. We optimize a dense feature-metric mean-squared error term in the inner loop and a four-corner homography error in the outer loop. The four-corner error is a simple measure that computes the L2 distance of four corners of the image after being transformed by the estimated and ground truth homography, as is used in \citep{detone16} as the output parameterization.

\vspace{-2.5mm}
\section{\baspacho: Batched Sparse Cholesky}\label{app:baspacho}
\vspace{-2.5mm}

In this section, we provide more details for our open-source novel \baspacho solver (\url{https://github.com/facebookresearch/baspacho}). \baspacho implements the \emph{supernodal} Cholesky algorithm~\citep{ng1993supernodal} to achieve state-of-the art performance by exploiting dense operations via \texttt{BLAS}/\texttt{cuBLAS}. The heuristics for clustering in the supernodal algorithm evaluate the trade-offs of fragmentation in sparse matrices against denser matrices with explicit zero-fill. These heuristics use a computation model that takes into account the architecture (batched/CPU/GPU) that can impact preference towards sparser or denser operations, and allows for further fine-tuning and customization. Apart from a minimal memory allocation needed for the symbolic factorization, \baspacho does not own any allocated memory allowing the user to fully manage memory arrays. This allow us to temporarily offload GPU arrays representing factorized matrix data to the CPU when necessary. Unlike existing solvers, \baspacho exposes lightweight random accessors that allows the user to read and write matrix blocks in the numeric factor data. This facilitates easy bookkeeping needed by optimization methods which often re-implement block-sparse matrix structures and convert between different matrix formats in order to invoke sparse solvers like CHOLMOD.

\begin{figure}
    \centering
    \includegraphics[width=\textwidth]{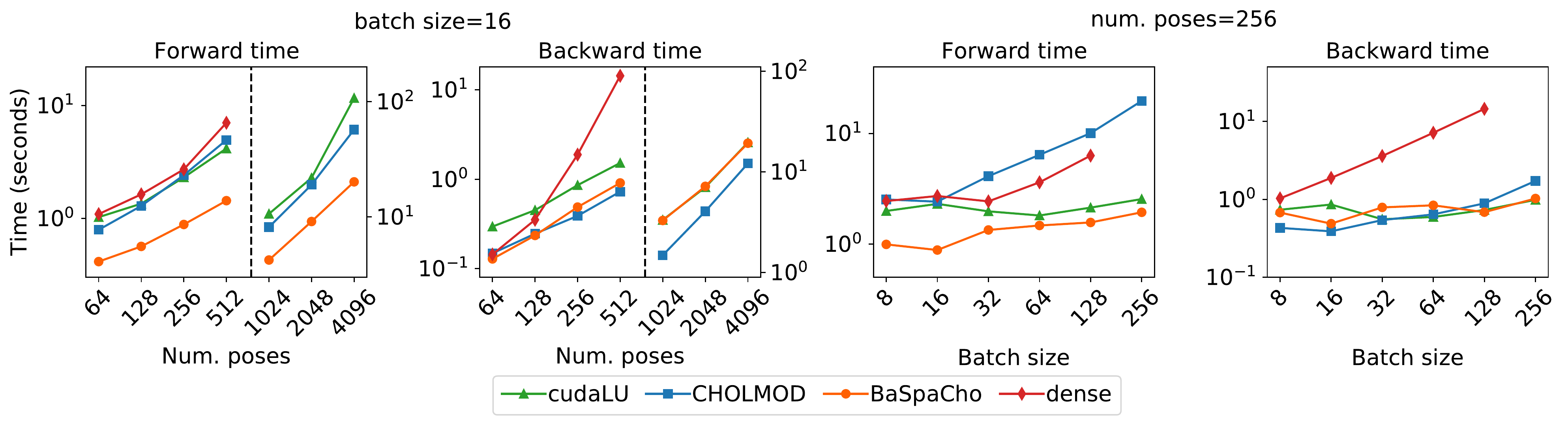}
    \vspace{-6mm}
    \caption{Forward/backward times of \theseus with sparse and dense solvers on different PGO problem scales.}
    \label{fig:profiling_small}
\end{figure}

\vspace{-2.5mm}
\section{Benchmark details and additional results}\label{app:bench}
\vspace{-2.5mm}

In this section, we present more profiling results for forward and backward pass of \theseus, using the same setup as \cref{sec:profiling}. For evaluation, we used the cube datasets of PGO  (see \cref{app:pgo}) with different numbers of poses, batch sizes, and levels of sparsity. In addition to the forward and backward times as a function of the numbers of poses and batch sizes reported in \cref{sec:profiling}, we further report and analyze the memory usage of \theseus with different solvers (\cusparse, \cholmodp, \baspacho and \dense) in various settings.

\vspace{-2.5mm}
\subsection{Forward and backward times with smaller batch size and number of poses}\label{app:profiling_small}
\vspace{-2.5mm}

We profile PGO using different linear solvers (\cusparse, \cholmodp, \baspacho and \dense) for fixed batch size of 16 and number of poses of 256. The setup is the same as that in \cref{sec:profiling} except that fixed batch size and number of poses are smaller. \cref{fig:profiling_small} shows the average time of a full forward and backward pass. Similar to \cref{fig:profiling} with larger fixed batch size and number of poses, the sparse solvers are faster than \dense. For the smallest problem considered (64 poses, 16 batch size), the total sum of average forward and backward times are 1.32s (\cusparse), 0.94s (\cholmodp), 0.54s (\baspacho), 1.23s (\dense) per batch. Increasing to 128 poses makes the sparse solvers noticeably faster than \dense: 1.98s (\dense) vs 1.80s (\cusparse), 1.53s (\cholmodp), and 0.80s (\baspacho). As the problem scale increases, the gap between the sparse and dense solvers widens: for the largest problem solvable with \dense (512 poses, 16 batch size) the average total times are 5.68s (\cusparse), 5.67s (\cholmodp), 2.34s (\baspacho), and 21.37s (\dense). The speedup over \dense is $\sim$3.7x for \cusparse and \cholmodp, and $\sim$9.1x for \baspacho.

\begin{figure}[t]
    \centering
    \vspace{0.5mm}
    \includegraphics[width=\textwidth]{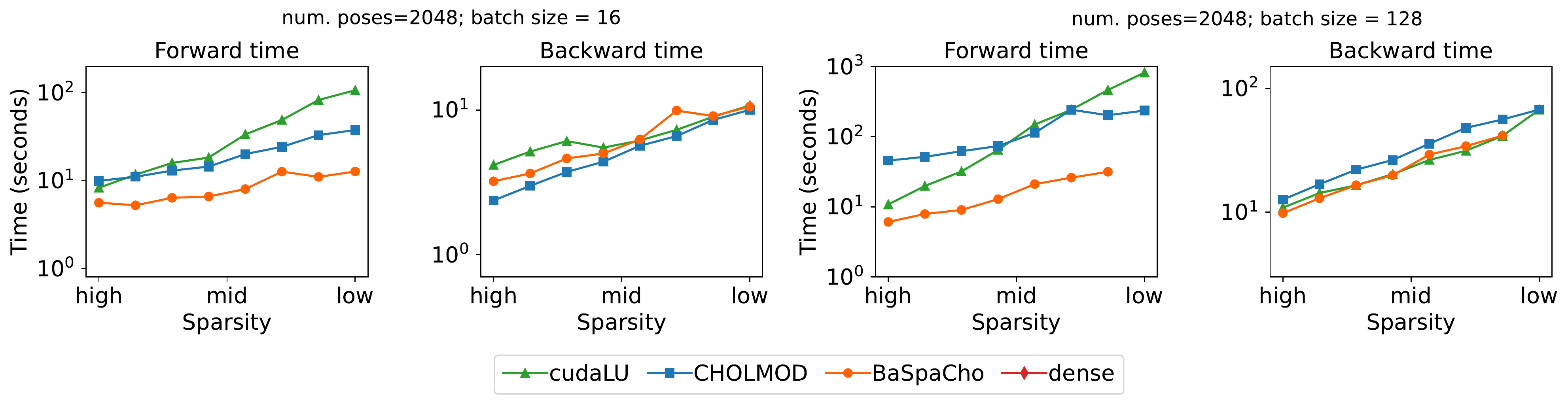}
    \caption{Forward and backward times of \theseus with sparse and dense solvers on PGO problems with 2048 poses and different levels of sparsity.}
    \label{fig:profiling-sparsity-time-large}
    \vspace{-1mm}
\end{figure}

\begin{figure}[t]
    \centering
    \includegraphics[width=\textwidth]{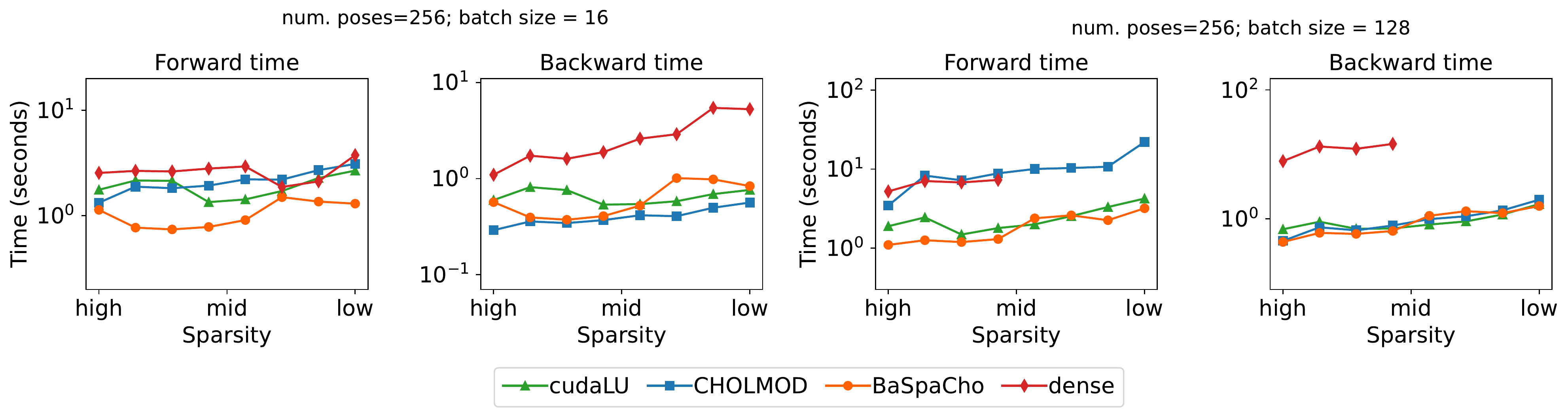}
    \caption{Forward and backward times of \theseus with sparse and dense solvers on PGO problems with 256 poses and different levels of sparsity.}
    \label{fig:profiling-sparsity-time-small}
    \vspace{-1mm}
\end{figure}

\vspace{-2.5mm}
\subsection{Forward and backward times with respect to sparsity}\label{app:sparsity_time}
\vspace{-2.5mm}

We study the forward and backward pass times of \theseus with sparse and dense solvers for different levels of sparsity using the synthetic Cube dataset. In PGO, loop closure probability represents how likely a pose has a loop closure edge connected to the other poses, and thus, greater loop closure probability yields a less sparse Hessian. We use loop closure probabilities from 0.05 to 0.40 in increments of 0.05, to indicate the level of sparsity for Cube datasets from high (0.05) to low (0.40).

The average forward and backward times of \theseus on PGO problems with different levels of sparsity for fixed numbers of poses are shown in \cref{fig:profiling-sparsity-time-large} for 2048 poses and in \cref{fig:profiling-sparsity-time-small} for 256 poses. In both figures, the left two plots are with 16 batch size and the right two plots are with 128 batch size. As expected, it takes more time in most cases for PGO problems with lower sparsity. Since \dense does not exploit the sparsity of optimization problems when solving the linear systems, forward pass of \dense takes almost the same amount of time regardless of the levels of sparsity. There is still some overhead for \dense as sparsity decreases, because more loop closure edges implies more cost function terms in the objective, so putting together the approximate Hessian is computationally more expensive.

\vspace{-2.5mm}
\subsection{Scalability of \theseus}
\vspace{-2.5mm}

\begin{wrapfigure}{r}{0.58\textwidth}
    \vspace{-12mm}
    \centering
    \includegraphics[width=0.58\textwidth]{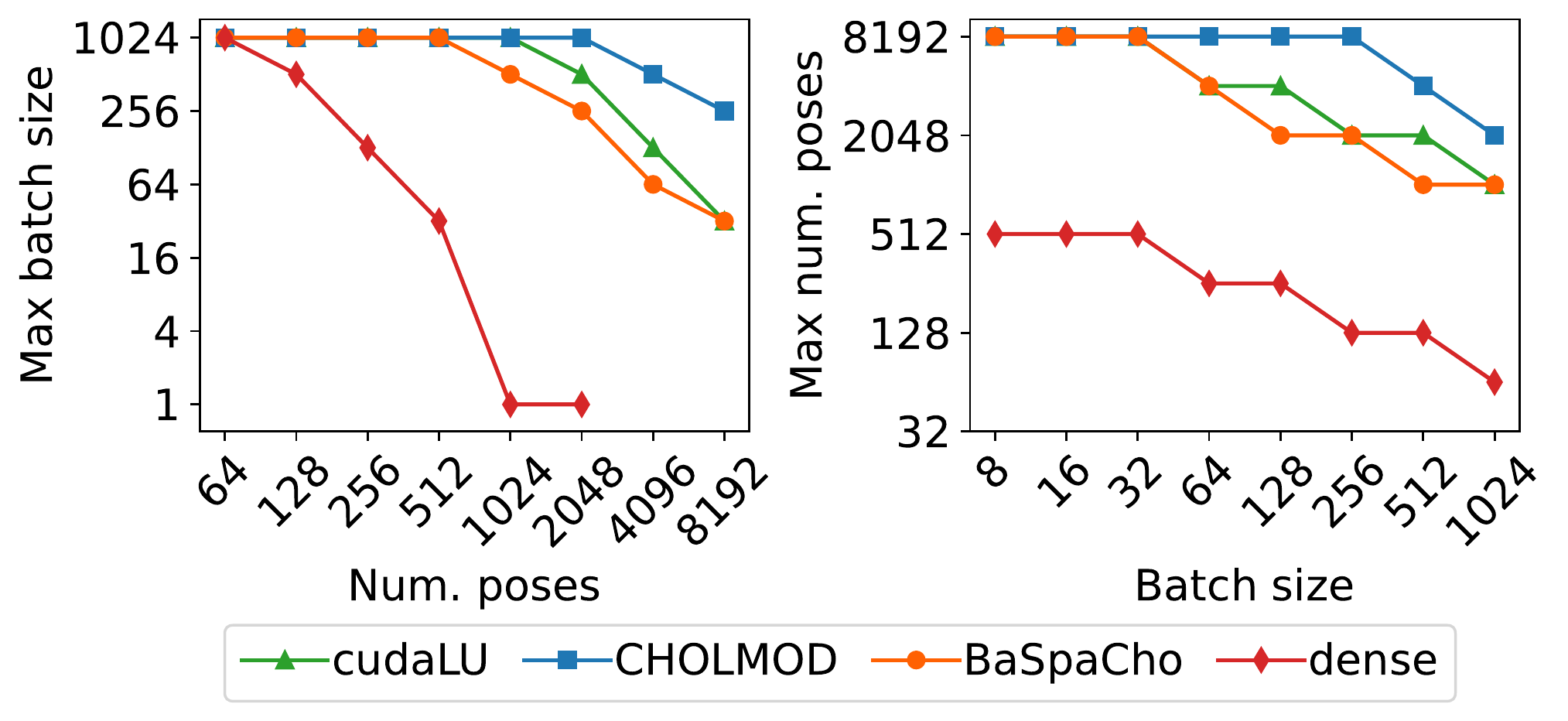}
    \caption{Largest PGO problems \theseus scales to for different numbers of poses and batch sizes.}
    \vspace{-1mm}
    \label{fig:profiling-size}
\end{wrapfigure}

In addition to forward and backward times in \cref{sec:profiling} and \cref{app:sparsity_time,app:profiling_small}, we analyze the scalability of \theseus with different linear solvers (\cusparse, \cholmodp, \baspacho and \dense) following a similar setup to evaluation in \cref{fig:profiling_small,fig:profiling}.

We profile PGO with various numbers of poses from 64 to 8192 and batch sizes from 8 to 1024 in increments of power of 2. \cref{fig:profiling-size} shows the maximum batch sizes solvable for given numbers of poses (left) and the maximum numbers of poses solvable for given batch sizes (right). In \cref{fig:profiling-size}, it can be seen that \dense only scales to small PGO problems due to the memory limitation and fails to solve any PGO problems with 4096 poses or more, even with batch size of 1. In contrast, \cusparse, \cholmodp and \baspacho successfully solve PGO problems with 8192 poses for a batch size of 32 (\cusparse, \baspacho) and 256 (\cholmodp).  As discussed in \cref{sec:profiling}, \cusparse and \baspacho require extra GPU memory to solve linear systems, whereas \cholmodp has all computation run on CPU, and thus can solve larger \dnls problems than \cusparse and \baspacho.

\vspace{-2mm}

\begin{wrapfigure}{r}{0.58\textwidth}
    \vspace{-5.5mm}
    \includegraphics[width=0.58\textwidth]{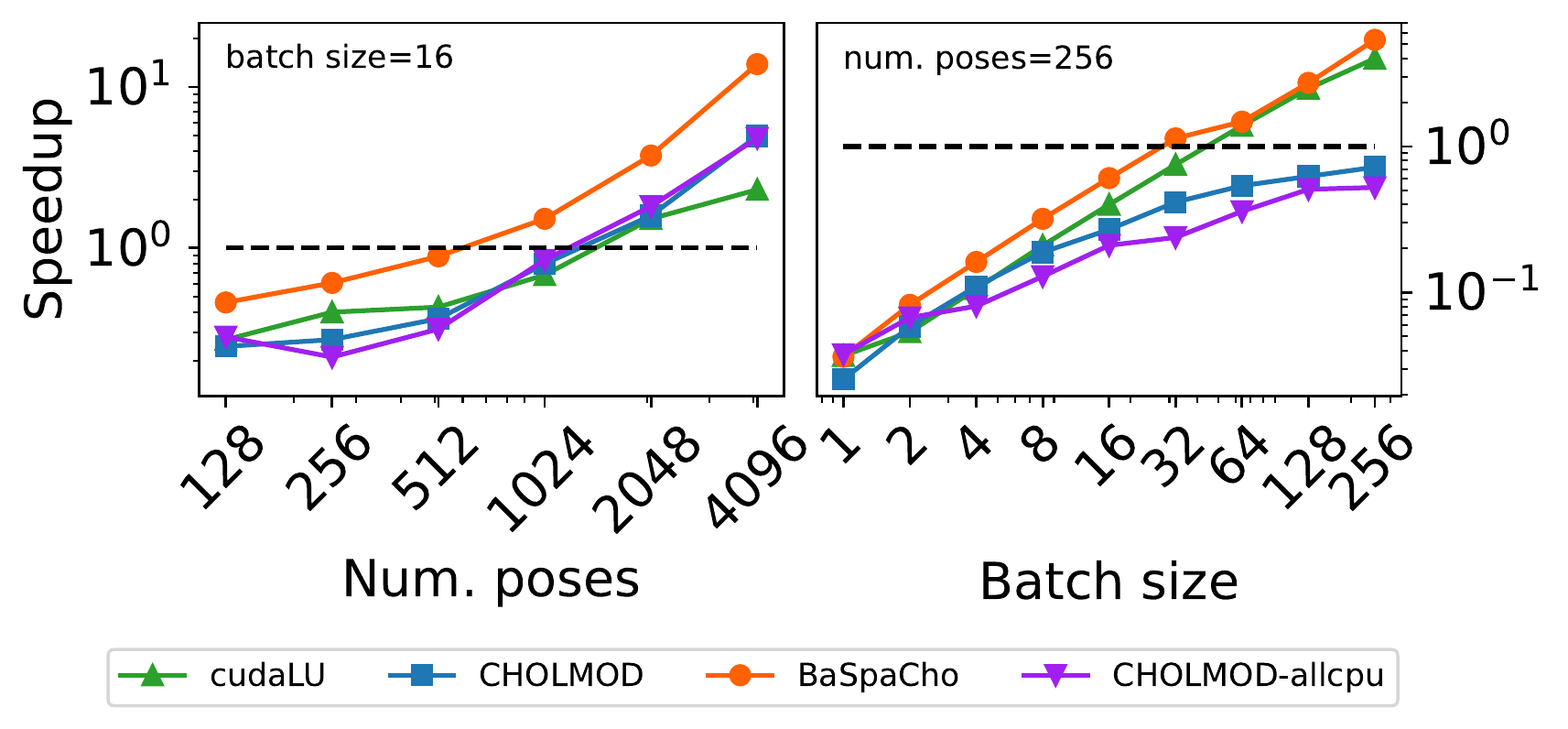}
    \vspace{-5.5mm}
    \caption{Speedup of \theseus (forward pass) over Ceres (black dashed) on different PGO problem scales.}
    \label{fig:ceres-comparison-small}
\end{wrapfigure}

\subsection{Comparison with \ceres on smaller batch size and number of poses}
\vspace{-2mm}

In addition to \cref{sec:ceres-comp}, we follow the same setup to compare \theseus as a stand-alone NLS optimizer with \ceres for PGO problems with a smaller fixed
batch size of 16 and number of poses of 256. \cref{fig:ceres-comparison-small} shows the speedup of \theseus compared to \ceres (black dotted line). Similar to \cref{fig:ceres-comparison}, \ceres is faster for PGO problems for small batch sizes and numbers of poses, and \theseus is faster as the problem scale increases.

\vspace{-2mm}
\section{Backward mode details, additional results, and derivations}\label{app:backward}

\vspace{-2mm}
\subsection{Experimental details}
\vspace{-2mm}

In~\Cref{sec:backward-experiments} we use the tactile state estimation example to evaluate the performance of different backward modes. As mentioned in~\Cref{app:tactile-problem}, the dataset consists of 63 trajectories of length 25, 56 of which we use for training and 7 for test. We use a batch size of 8 and train for 100 epochs, resulting in 700 batches for averaging time and memory results. For the inner loop, we use Gauss-Newton with a step size of 0.05; in the test set we run the inner loop for 50 iterations, regardless of the number used during training. For the outer loop, we use the Adam optimizer with a learning rate of $10^{-4}$, decayed exponentially by a factor of 0.98 after every epoch. 

\vspace{-2mm}
\subsection{Additional results}
\vspace{-2mm}

\Cref{fig:backward-mem-losses} (left) shows the peak memory consumption during the forward pass. We observe the same trend from the backward pass (\Cref{fig:time-and-mem-tactile}, center right), where \unroll's memory consumption increases linearly with the number of inner loop iterations, while for the other methods it remains constant. \implicit has the lowest peak memory requirement ($\sim$22MBs), followed by \dlm ($\sim$29MBs).

\Cref{fig:backward-mem-losses} (center, right) also shows training curves for all methods. We observe that, despite higher performance in the test set, \implicit is more unstable during training and oscillates between low and high values; this suggests that careful use of early stopping and hyperparameter tuning might be required when using \implicit. The other methods are more stable, with the two truncated methods achieving the lowest training loss after \implicit. \Cref{fig:backward-mem-losses} (right) shows that \unroll's performance degradation, relative to other methods, with increasing number of inner loop iterations (also see~\Cref{fig:time-and-mem-tactile}, right) is not just a generalization issue, but also happens during training. This suggest possible numerical issues from unrolling gradients over a high number of optimization steps, as observed in prior work.

\begin{figure}[h]
    \centering 
    \includegraphics[width=\textwidth]{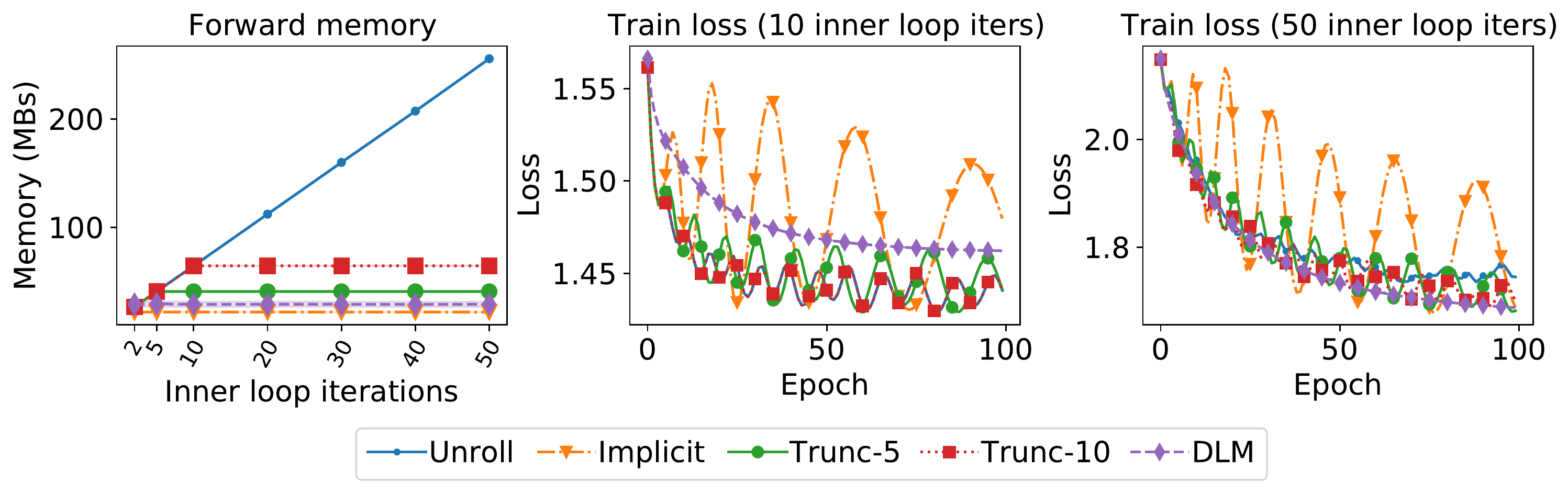}
    \caption{Additional results for backward modes comparison in tactile state estimation problem. \textbf{Left:} Memory consumption in forward pass. \textbf{Center:} Training loss when using 10 inner loop iterations. \textbf{Right:} Training loss when using 50 inner loop iterations.}
    \label{fig:backward-mem-losses}
\end{figure}

\vspace{-2mm}
\subsection{Backward modes summary and limitations}
\vspace{-2mm}

\Cref{fig:backward_modes_vis} visualizes the backward modes
and \cref{tab:backward_modes_compatibility}  contrasts their limitations.
The table shows that all four modes can be used when learning parameters
for cost functions or cost weights.
However, unlike other approaches, \implicit cannot be used
when learning initial values for the optimization variables,
$\theta_{\text{init}}$. Another limitation of \implicit is that the resulting
gradients might be inaccurate in problems where it is not feasible to
find the optimal solution to the inner optimization problem; other
methods don't experience this limitation, since they compute gradients
around the approximate solution found. On the other hand, both \unroll
and \trunc could potentially experience vanishing or exploding
gradient issues when the number of iterations to backpropagate through
is large, a limitation that is not shared by \implicit and
\dlm. Finally, a limitation of \dlm is that $\epsilon$ needs to be
tuned (see~\cref{eq:dlm_theta_direct}), which can greatly affect
performance. Likewise, the number of backward iterations for \trunc
may also require some tuning.

\begin{table}[t]
    \scriptsize
    \vspace{0.5cm}
	\centering
	\setlength{\tabcolsep}{.5em}
	\begin{tabular}{c| c c c c c c}
		\toprule
		& \thead{Cannot be used for \\ learning $c_i$ or $w_i$} & \thead{Cannot be used \\ for learning $\theta_{\text{init}}$}  & \thead{Requires accurate \\ $\theta^*$ solution} & \thead{Possible vanishing or \\ exploding gradients}  & \thead{Requires \\ tuning} & \thead{Compute and \\ memory usage} \\
		\midrule
		\unroll &  & &  & \cmark & & high \\
		\trunc &  &  &  & \cmark & \cmark & medium\\
		\implicit & & \cmark & \cmark & & & low \\
		\dlm &  &  &  &  & \cmark & low \\
		\bottomrule
	\end{tabular}
	\vspace{0.5em}
	\caption{Backward modes summary and limitations.}
	\vspace{-6mm}
	\label{tab:backward_modes_compatibility}
\end{table}

\begin{figure}[t]
  \centering
  \unroll \\
  \newcommand{\dottts}{\hspace{1.4mm}\phantom{\ldots\ldots..}\ldots\phantom{.\ldots\ldots}\hspace{1.4mm}}
  \begin{tikzpicture}
    \matrix (m) [
    matrix of math nodes,row sep=2em,column sep=1em,
    minimum width=2em,nodes={anchor=center}
    ] {
      \theta_0 & \theta_{1} &
      \dottts & \theta_{K} & \theta^\star(\phi) & \gL(\theta^\star(\phi)) \\
      \; & \; & \; & \; & \; & \; \\};
    \node at (-0.1,0.) () {\color{backward_arrow_color}\ldots};
    \path[-stealth]
    (m-1-1) edge node {} (m-1-2)
    (m-1-2) edge node {} (m-1-3)
    (m-1-3) edge node {} (m-1-4)
    (m-1-4) edge node {} (m-1-5)
    (m-1-5) edge node {} (m-1-6)
    (m-1-6) edge[draw=backward_arrow_color,out=-155,in=-30] node[below] {} (m-1-5)
    (m-1-6) edge[draw=backward_arrow_color,out=-155,in=-30] node[below] {} (m-1-4)
    (m-1-6) edge[draw=backward_arrow_color,out=-155,in=-15] node {} (m-1-2)
    (m-1-6) edge[draw=backward_arrow_color,out=-155,in=-15] node {} (m-1-1)
    ;
    \draw (current bounding box.north east) rectangle ([yshift=5mm]current bounding box.south west);
  \end{tikzpicture} \\[-3mm]

  \trunc \\
  \begin{tikzpicture}
    \matrix (m) [
    matrix of math nodes,row sep=2em,column sep=1em,
    minimum width=2em,nodes={anchor=center}
    ] {
      \theta_0 & \theta_1 &
      \ldots & \theta_{K-H} & \ldots &
      \theta_K &  \theta^\star(\phi) & \gL(\theta^\star(\phi)) \\
      \; & \; & \; & \; & \; & \; \\};
    \node at (1.2,0.1) () {\color{backward_arrow_color}\ldots};
    \path[-stealth]
    (m-1-1) edge node {} (m-1-2)
    (m-1-2) edge node {} (m-1-3)
    (m-1-3) edge node {} (m-1-4)
    (m-1-4) edge node {} (m-1-5)
    (m-1-5) edge node {} (m-1-6)
    (m-1-6) edge node {} (m-1-7)
    (m-1-7) edge node {} (m-1-8)
    (m-1-8) edge[draw=backward_arrow_color,out=-155,in=-30] node[below] {} (m-1-7)
    (m-1-8) edge[draw=backward_arrow_color,out=-155,in=-30] node[below] {} (m-1-6)
    (m-1-8) edge[draw=backward_arrow_color,out=-155,in=-15] node {} (m-1-4)
    ;
    \draw (current bounding box.north east) rectangle ([yshift=6mm]current bounding box.south west);
  \end{tikzpicture} \\[-3mm]

  \implicit \\
  \begin{tikzpicture}
    \matrix (m) [
    matrix of math nodes,row sep=2em,column sep=1em,
    minimum width=2em,nodes={anchor=center}
    ] {
      \theta_0 & \theta_1 &
      \dottts & \theta_K & \theta^\star(\phi) & \gL(\theta^\star(\phi)) \\
      \; & \; & \; & \; \\};
    \path[-stealth]
    (m-1-1) edge node {} (m-1-2)
    (m-1-2) edge node {} (m-1-3)
    (m-1-3) edge node {} (m-1-4)
    (m-1-4) edge node {} (m-1-5)
    (m-1-5) edge node {} (m-1-6)
    (m-1-6) edge[draw=backward_arrow_color,out=-155,in=-30] node[below] {} (m-1-5)
    ;
    \draw (current bounding box.north east) rectangle ([yshift=8mm]current bounding box.south west);
  \end{tikzpicture} \\[-3mm]

  \dlm \\
  \begin{tikzpicture}
    \matrix (m) [
    matrix of math nodes,row sep=1em,column sep=1em,
    minimum width=2em,nodes={anchor=center}
    ] {
      \theta_0 & \theta_1 &
      \dottts & \theta_K & \theta^\star(\phi) & \gL(\theta^\star(\phi)) \\
      \; & \; & \; & \; & \theta_{\rm direct}(\phi) \hspace*{-6mm}  \\};
    \path[-stealth]
    (m-1-1) edge node {} (m-1-2)
    (m-1-2) edge node {} (m-1-3)
    (m-1-3) edge node {} (m-1-4)
    (m-1-4) edge node {} (m-1-5)
    (m-1-5) edge node {} (m-1-6)
    (m-1-6) edge[draw=backward_arrow_color] node[above] {} (m-2-5)
    (m-2-5) edge[draw=backward_arrow_color] node[above] {} (m-1-5)
    ;
    \draw (current bounding box.north east) rectangle ([yshift=1mm]current bounding box.south west);
  \end{tikzpicture} \\
  \caption{Illustration of the dependencies of the backward modes
    for computing $\nabla_\phi \gL(\theta^\star)$.}
  \label{fig:backward_modes_vis}
  \vspace{0.5mm}
\end{figure}
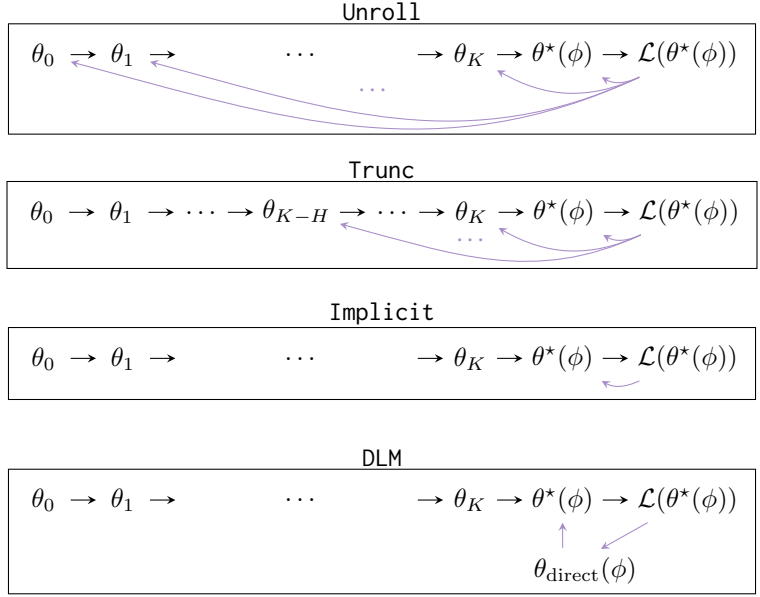


\subsection{Derivations for backward modes}

\vspace{-1mm}
\subsubsection{Implicit function theorem}
\vspace{-2mm}

For adjoint differentiation, we make use of the implicit function
theorem, which is originally from \citet{dini1878analisi},
and presented in \citet[Theorem 1B.1]{dontchev2009implicit} as:
\begin{theorem}[Dini's implicit function theorem]
  \label{thm:ift}
  Let the roots of $g(\theta; \phi)$ define an implicit
  mapping $\Theta^\star(\phi)$ given by $\Theta^\star(\phi)\defeq\{\theta \mid g(\theta;\phi)=0\}$,
  where $\theta\in\R^m$, $\phi\in\R^n$, and
  $g: \R^m\times\R^n\rightarrow\R^m$.
  Let $g$ be continuously differentiable in a neighborhood of $(\bar \theta, \bar \phi)$
  such that $g(\bar \theta; \bar \phi)=0$, and let the Jacobian of $g$
  with respect to $\theta$ at $(\bar \theta, \bar \phi)$,
  \ie $\D_\theta g(\bar \theta; \bar \phi)$, be non-singular.
  Then $\Theta^\star$ has a single-valued localization $\theta^\star$
  around $\bar \phi$ for $\bar \theta$ which is continuously differentiable
  in a neighborhood $Q$ of $\bar \phi$ with Jacobian satisfying
  \begin{equation}
    \D_\phi \theta^\star(\tilde \phi) = -\D_\theta^{-1} g(\theta^\star(\tilde \phi); \tilde \phi) \D_\phi g(\theta^\star(\tilde \phi); \tilde \phi)
    \qquad \mathrm{for\ every}\; \tilde \phi\in Q.
    \label{eq:implicit-function-theorem}
  \end{equation}
\end{theorem}

\vspace{-2mm}
\subsubsection{Proof of \texorpdfstring{\cref{prop:newton-ift}}{Prop. 1}}
\vspace{-2mm}

\begin{proof}
\looseness=-1
  Let $\bar \phi$ be a hyper-parameter resulting in a unique $\theta^\star(\bar \phi)$
  and recall that the implicit mapping for \cref{eq:nls} is defined by
  $g(\theta; \phi)\defeq \nabla_\theta S(\theta, \phi)$ and is
  zero at the optimal parameters, \ie $g(\theta^\star(\bar \phi), \bar \phi)=0$.
  Let $h(\theta; \phi):=\theta-[\nabla^2_\theta S(\theta; \phi)]^{-1}_{\rm stop}\nabla_\theta S(\theta; \phi)$ be the Newton iteration
  where $[\cdot]_{\rm stop}$ is a function that zeros the derivative.
  Differentiating $h$, which can be done using automatic differentiation
  on a Newton step, results in the implicit derivative
  \cref{eq:implicit-function-theorem}:
  \begin{equation}
    \label{eq:diff-newton-ift}
    \begin{aligned}
    \D_\phi h(\theta^\star(\bar \phi); \bar \phi) &=
    \D_\phi \left( \theta-\left[\nabla^2_\theta
          S(\theta^\star(\bar \phi); \bar \phi)\right]^{-1}_{\rm stop}
        \nabla_\theta S(\theta^\star(\bar\phi); \bar\phi) \right) \\
    &= -\left[\nabla^2_\theta S(\theta^\star(\bar \phi); \bar \phi)\right]^{-1}_{\rm stop}
        \D_\phi \nabla_\theta S(\theta^\star(\bar\phi); \bar\phi) \\
    &= -\D_\theta^{-1} g(\theta^\star(\bar \phi); \bar \phi)
        \D_\phi g(\theta^\star(\bar\phi); \bar\phi)
    \end{aligned}
  \end{equation}
\end{proof}

\vspace{-2mm}
\subsubsection{Direct loss minimization for use in \theseus}
\vspace{-2mm}

Originally, DLM gradient for non-linear objective functions~\citep{song2016training} can be expressed as
\begin{equation}\label{eq:dlm}
    \nabla_\phi L = \lim_{\varepsilon \rightarrow 0} g_\text{DLM}^\varepsilon, \; \text{ where } \;
    g_\text{DLM}^\varepsilon \triangleq \frac{1}{\varepsilon} \left[ \frac{\partial}{\partial \phi} S(\theta^*; \phi) - \frac{\partial}{\partial \phi} S(\theta_\text{direct}; \phi) \right]
\end{equation}
where
\begin{equation}
    \theta^* = \argmin_{\hat\theta} S(\hat\theta; \phi),
    \quad\quad\text{ and }\quad\quad
    \theta_\text{direct} = \argmin_{\hat\theta} S(\hat\theta; \phi) - \varepsilon L(\hat\theta).
\end{equation}
However, this dependence on the loss function fits poorly in a reverse-mode automatic differentiation framework like \pytorch. Instead, we can construct an equivalent formulation by noting that in continuous space, we can first linearize the loss function around the current solution $\theta^*$, 
\begin{equation}
    \hat{L}(\theta) = L(\theta^*) + \nabla_{\theta} L(\theta^*)(\theta - \theta^*)
\end{equation}
Let $v = \nabla_{\theta} L(\theta^*)$, then the perturbed solution becomes
\begin{equation}
    \theta_\text{direct} 
    = \argmin_{\hat\theta} S(\hat\theta; \phi) - \varepsilon \left(L(\theta^*) + v^T (\hat\theta - \theta^*)\right)
    = \argmin_{\hat\theta} S(\hat\theta; \phi) - \varepsilon v^T \hat\theta.
\end{equation}
Plugging this back into \Cref{eq:dlm}, we see that this is an algorithm which takes in a gradient vector $v$ and computes an approximation to the vector-Jacobian product $\nabla_{\phi} L(\theta^*) = v \frac{\partial \theta^*}{\partial \phi}$.

As \theseus is designed to solve optimization problems where $S$ is expressed as sum of squares, it cannot readily handle solving $\theta_{direct}$ as this requires adding a linear term to the objective. Instead, let us consider the following ``completing the square'' approach:
\begin{equation}
\arg\min_{\hat\theta} \| \varepsilon \hat\theta \|^2 - \varepsilon v^T \theta
= \arg\min_{\hat\theta} \varepsilon^2 \hat\theta^T \hat\theta - \varepsilon v^T \hat\theta + \left(\tfrac{1}{2} v\right)^T \left(\tfrac{1}{2} v\right)
= \arg\min_{\hat\theta} \left\| \varepsilon \hat\theta - \tfrac{1}{2} v \right\|^2
\end{equation}
We can thus add this extra term and let
\begin{equation}
    \theta_\text{direct} = \argmin_{\hat\theta} S(\hat\theta; \phi) + \left\| \varepsilon \hat\theta - \tfrac{1}{2} v \right\|^2
    \label{eq:dlm_theta_direct}
\end{equation}
This adds a small bias to the gradient due to the addition of $\| \varepsilon \hat\theta \|^2$ but when $\varepsilon$ is small it shouldn't be problematic. In practice, we solve for $\theta_\text{direct}$ by starting from $\theta^*$ and using just one iteration of Gauss-Newton.

}\fi}\fi

\bibliographystyle{unsrtnat}
\bibliography{theseus}

\ifARXIV{}\fi

\ifARXIV{}\fi

\end{document}